\documentclass[10pt,twocolumn,letterpaper]{article}

\usepackage{cvpr}              

\usepackage{url}
\usepackage{graphicx}

\usepackage{booktabs}       
\usepackage{amsfonts}       
\usepackage{nicefrac}       
\usepackage{microtype}      
\usepackage[table,xcdraw,dvipsnames]{xcolor}         
\usepackage[most]{tcolorbox}
\usepackage{algorithm}
\usepackage{algpseudocode}
\usepackage{multirow}
\usepackage{amsmath}
\usepackage{pifont}
\usepackage{wrapfig}

\definecolor{cvprblue}{rgb}{0.21,0.49,0.74}
\usepackage[pagebackref,breaklinks,colorlinks,allcolors=cvprblue]{hyperref}

\title{HiconAgent: History Context-aware Policy Optimization for GUI Agents}

\author{
Xurui Zhou$^{1}$,Gongwei Chen$^{1}$,  Yuquan Xie$^{1}$, Zaijing Li$^{1}$,
Kaiwen Zhou$^{3}$,\\
Shuai Wang$^{3}$, Shuo Yang$^{1}$, Zhuotao Tian$^{1}$,
Rui Shao$^{1,2}$\footnotemark[2]
\\
    $^{1}$Harbin Institute of Technology, Shenzhen\quad
    $^{2}$Shenzhen Loop Area Institute\quad
    $^{3}$Huawei Noah's Ark Lab\\
    \texttt{\normalsize{zhouxurui1314@gmail.com}}\hspace{0.5cm} \texttt{\normalsize{chengongwei@hit.edu.cn}}\hspace{0.5cm}\texttt{\normalsize{shaorui@hit.edu.cn}}
    \\
    \texttt{\normalsize{\url{https://github.com/JiuTian-VL/HiconAgent}}}
}

\begin{document}
\maketitle
\begin{abstract}
Graphical User Interface (GUI) agents require effective utilization of historical context to perform sequential navigation tasks. While incorporating past actions and observations can significantly improve decision-making, naively using full history leads to excessive computational overhead and potential distraction from irrelevant information. In this work, we introduce \textbf{HiconAgent}, a GUI agent trained with \textbf{History Context-aware Policy Optimization (HCPO)} for effective and efficient utilization of historical information. HCPO explicitly optimizes history usage in both sampling and policy updates by integrating two complementary components: \textbf{(1) Dynamic Context Sampling (DCS)} presents the agent with variable-length histories during sampling, enabling adaptive use of the most relevant historical context to improve sequential decision quality; \textbf{(2) Anchor-guided History Compression (AHC)} refines the policy update phase via a dual-branch optimization strategy, where the compressed branch drops history observations while keeping history actions as information flow anchors. The compressed and uncompressed branches are coupled through a history-enhanced alignment loss to enforce consistent history usage, achieving efficiency with minimal performance degradation. Extensive experiments on mainstream GUI navigation benchmarks demonstrate the strong performance of our model. Despite its smaller size, HiconAgent-3B outperforms GUI-R1-7B by \textbf{+8.46}\% grounding and \textbf{+11.32}\% step successful rate on GUI-Odyssey, while achieving comparable results on AndroidControl and AITW, with up to \textbf{2.47× computational speedup} and \textbf{60\% FLOPs reduction}. 
\end{abstract}
\vspace{-5mm}

\section{Introduction}
Multimodal Large Language Model (MLLM)-based GUI agents~~\cite{cheng_seeclick_2024, wu_os-atlas_2024} have recently achieved strong performance in grounding and navigation tasks by leveraging textual instructions, visual observations, and historical trajectories. Among various training strategies, reinforcement learning (RL)~\cite{luong_reft_2024,liu_visual-rft_2025,tan_reason-rft_2025} has demonstrated strong effectiveness by allowing GUI agents to directly optimize task-oriented objectives such as grounding accuracy and successful rate. Compared to supervised learning approaches~\cite{xu_aguvis_2024,gou2024uground}, RL-based methods~\cite{lu_ui-r1_2025,luo2025gui} significantly improve decision quality, robustness and generalization. As a result, RL has become a mainstream paradigm for training advanced GUI agents.

\begin{figure}
    \centering
    \includegraphics[width=1.0\linewidth]{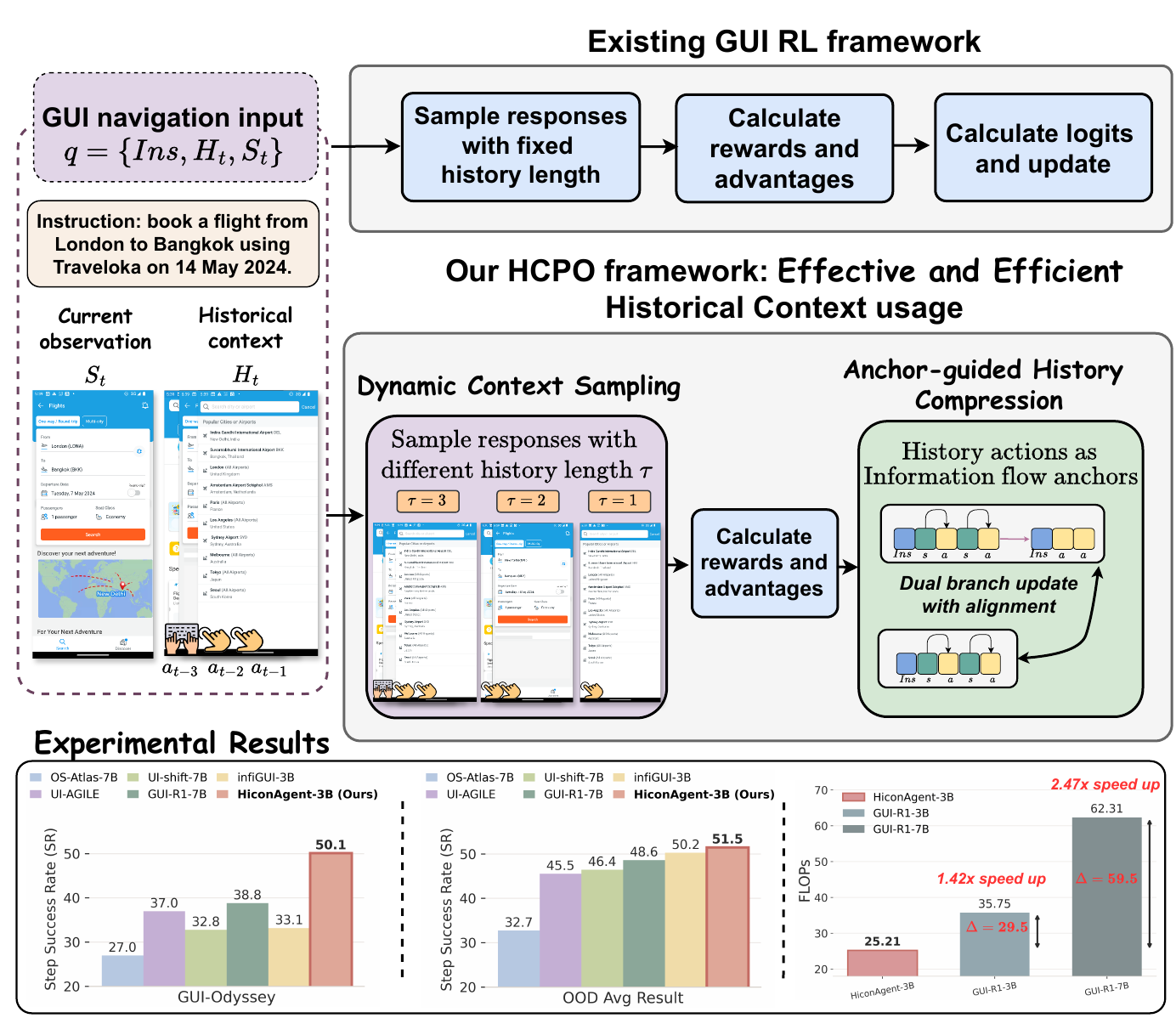}
    \vspace{-5mm}
    \caption{Comparison of existing GUI RL framework with our HCPO framework. HCPO jointly improves the sampling and update phases of training by integrating Dynamic Context Sampling \textbf{(DCS)} and Anchor-guided History Compression \textbf{(AHC)}.}
    \label{fig:teaser}
    \vspace{-6mm}
\end{figure}

Despite these advances, the role of \textbf{\textit{history usage}} in RL-based GUI agents remains largely underexplored. Most prior works ~\cite{lu_ui-r1_2025,luo2025gui,lian2025ui} adopt a simplified design in which history observations (past screenshots) are omitted, and only history actions are included as the input context. While this choice reduces memory and computational cost, it discards rich visual cues from past observations that are often essential for resolving ambiguous instructions, grounding visually similar elements, and maintaining temporal consistency across steps~\cite{UI-hawk}. Conversely, naively incorporating complete history, including both past actions and observations, substantially increases computational overhead due to the quadratic complexity of attention mechanisms and the large number of visual tokens from high-resolution screenshots. This trade-off between decision quality and efficiency motivates the development of methods that can effectively retain the most informative parts of historical context while mitigating redundancy.

To this end, we propose History Context-aware Policy Optimization (\textbf{HCPO}), a training framework designed to improve both the effectiveness and efficiency of history usage in GUI agents. As illustrated in Figure \ref{fig:teaser}, HCPO improves both the sampling and update phases of existing GUI RL framework through two complementary components: Dynamic Context Sampling (\textbf{DCS}) and Anchor-guided History Compression (\textbf{AHC}). 

\textit{\textbf{In the sampling phase}}, DCS addresses the variability of history dependence across decision steps. Different from conventional RL approaches that use a fixed-length history, DCS samples multiple history variants for each rollout using the exponential-biased distribution, encouraging the model to adaptively utilize the most relevant context. Our sampling distribution is motivated by empirical observations, ensuring stable learning and avoiding the degeneration observed with naive uniform sampling. \textit{\textbf{In the update phase}}, AHC jointly optimizes compressed and uncompressed branches to enhance policy updates under history compression. The compressed branch drops history observations after early fusion, retaining only action tokens as anchors to preserve essential decision signals. It is jointly optimized with the uncompressed branch through a history-enhanced alignment loss. This compression strategy is guided by our empirical layer-wise token-drop analysis.
\textbf{Our contributions} can be summarized as follows:
\begin{itemize}
    \item We conduct a \textbf{comprehensive empirical analysis} on history usage in GUI agents. Our findings show that different tasks and decision steps prefer different history lengths and history actions act as critical anchors for visual information flow. These findings reveal important inefficiencies in existing designs and directly motivate our method. 
    \item We propose History Context-aware Policy Optimization (\textbf{HCPO}), a novel reinforcement fine-tuning framework that combines Dynamic Context Sampling and Anchor-guided History Compression. Together, they enable agents to learn adaptive history usage while reducing redundancy and preserving decision quality.
    \item We validate our method on three GUI navigation benchmarks: GUI-Odyssey, AndroidControl, and AITW. HiconAgent-3B consistently outperforms existing reinforcement learning based agents. It outperforms larger 7B agents such as GUI-R1-7B by \textbf{+11.32\%} in step successful rate on GUI-Odyssey, while achieving \textbf{2.47×}  computational speedup and a \textbf{60\%} reduction in FLOPs.
\end{itemize}

\section{Related Work}
\subsection{GUI Agents and History Utilization}
GUI agents aim to complete high-level tasks by interacting with graphical user interfaces through sequences of low-level actions such as clicking and typing. Recent advances in multimodal large language models~\citep{chen2024lion,shen2024mome,zhang2025falcon,li2025lion, li2024optimus, achiam2023gpt,hurst2024gpt, li2025semanticvla, zhu2025h, li2025cogvla, shao2019multi, shao2023detecting, shao2024detecting, shao2017deep, lyu2025puma, zou2025enhancing, chen2020scene, jiang2025dive, li2024enhancing, NEURIPS2021_27d52bcb, li2025optimus2} have rapidly expanded their capability to integrate and reason over diverse input modalities. This progress has enabled GUI agents to achieve significant improvements in multi-step navigation tasks~\citep{cheng_seeclick_2024, wu_os-atlas_2024,wang_mobile-agent_2024, xie2025gui, chen2024spa, lyu2026personalalign, chen2019transferability, chen2025enhancing, xie2025mirage, li2025optimus}.

Although historical information, including past observations and actions, is essential for GUI agents to understand user instructions and make decisions, the issue of redundancy in long histories remains a persistent challenge. Existing studies~\citep{zhang_ui-hawk_2024, lu_gui_2024, chen2025less} under supervised fine-tuning show that adding past actions improves performance with minimal input cost. In contrast, incorporating full visual history brings larger gains but incurs significantly higher computation, especially in long-horizon or high-resolution scenarios. This trade-off highlights the inefficiency of naively using full history and motivates more selective or compressed representations of historical context.

\subsection{Rule-based Reinforcement Learning}
Large language models have shown remarkable improvements in reasoning-intensive tasks such as mathematics and programming when optimized with reinforcement learning. This reinforcement fine-tuning paradigm is now being actively explored in the multimodal domain. In particular, Group Relative Policy Optimization (GRPO)~\citep{shao_deepseekmath_2024, deepseek-ai_deepseek-r1_2025} has emerged as an effective alternative to conventional methods such as PPO~\citep{Schulman2017}, especially for training multimodal models. It simplifies the training pipeline by evaluating responses using relative, normalized rewards computed within each sample group.

In the GUI domain, rule-based RL typically uses exact match or distance-based criteria for action types, coordinates, or textual inputs, providing fine-grained rewards that guide the agent toward correct execution~\citep{luo2025gui, luo2025navimaster}. Although existing RL approaches have achieved notable performance gains, there has been little to no discussion on \emph{effective and efficient history utilization}, and existing methods typically overlook strategies for balancing the trade-off between computational cost and the retention of critical historical context. Our method extends rule-based RL with \emph{history-aware policy optimization} to improve decision quality while reducing computational overhead.

\begin{figure*}[t]
    \centering
    \includegraphics[width=0.49\linewidth]{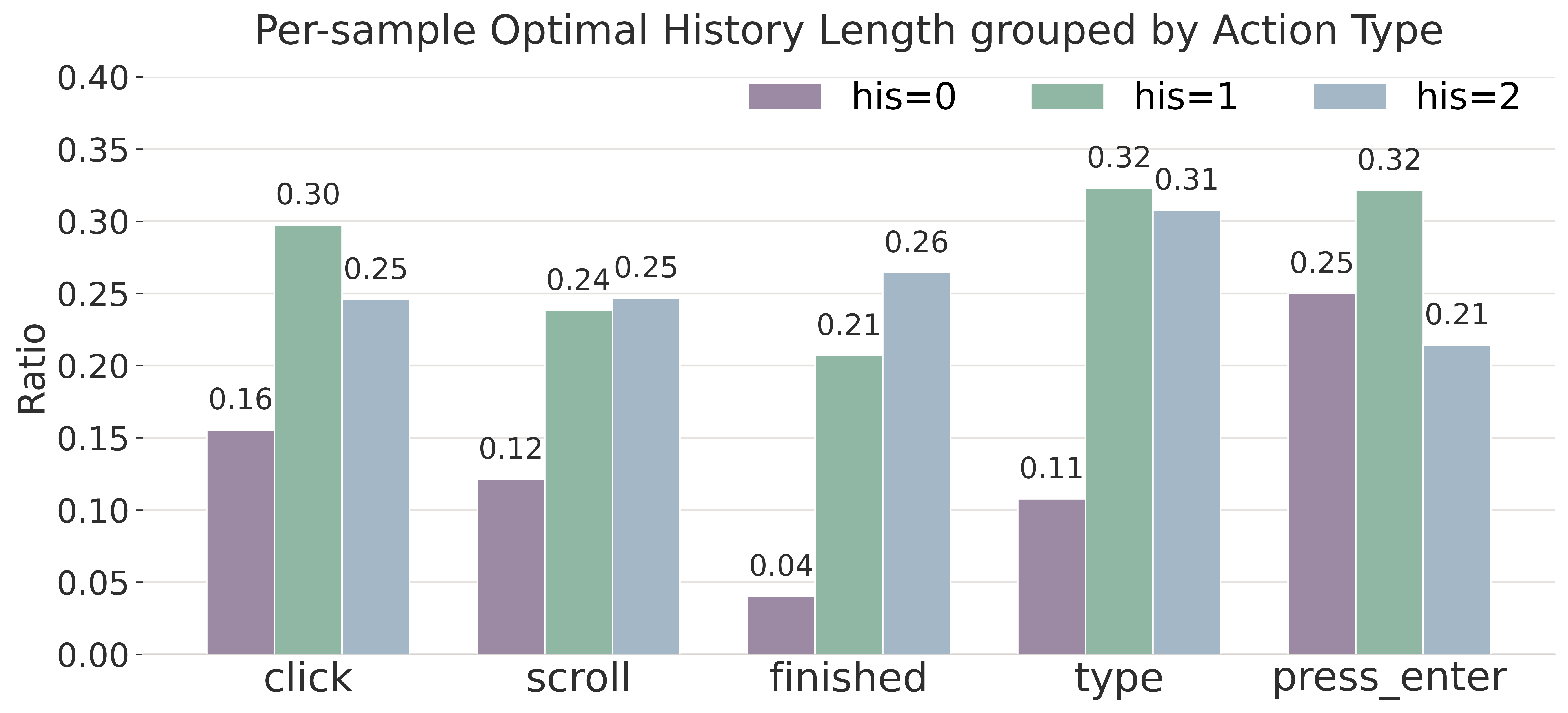}
    \hfill
    \includegraphics[width=0.49\linewidth]{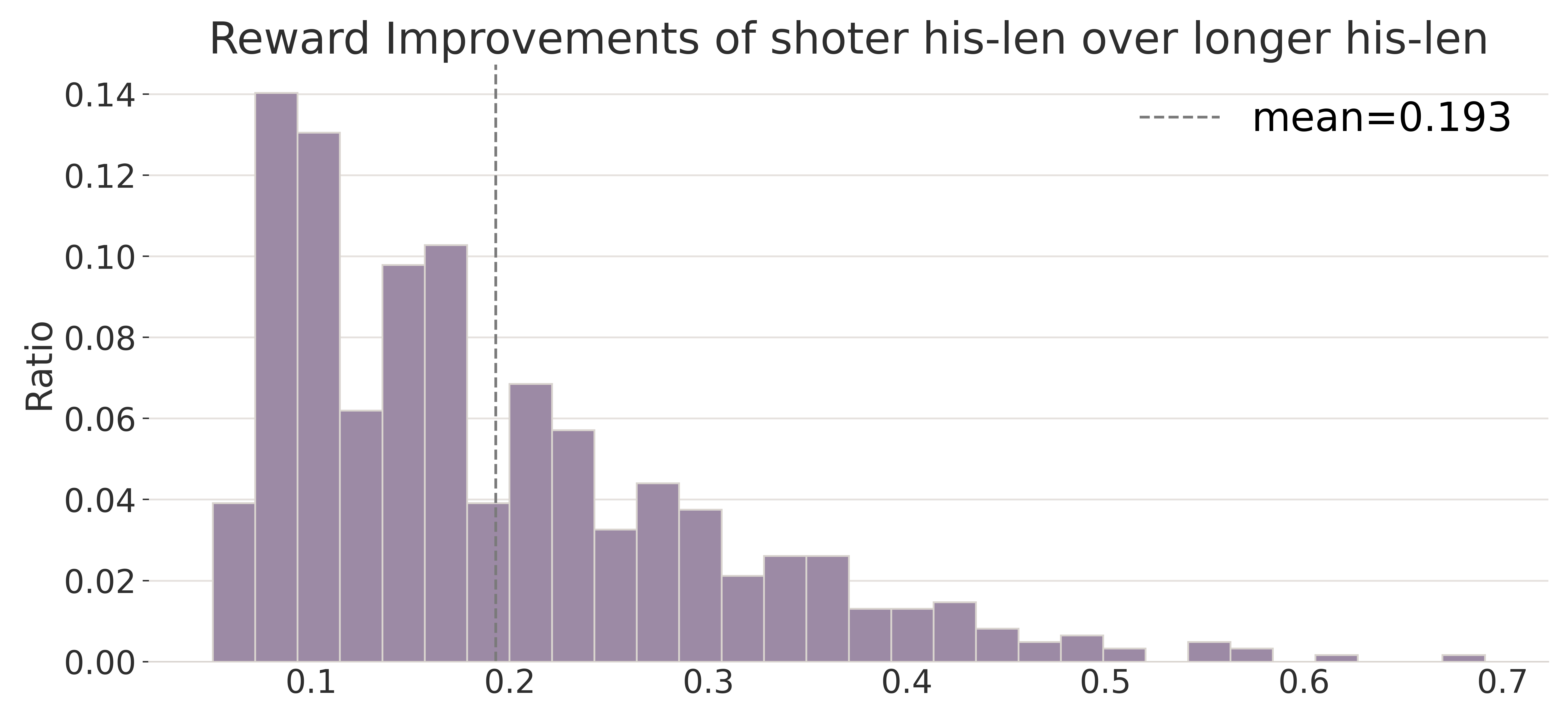}\
    \vspace{-1mm}
    \caption{\textbf{Different samples prefer different history lengths.}
    \emph{Left}: For each sample we evaluate a set of different history lengths~$\tau$ and take the $\tau$ that yields the highest mean reward. The preferred $\tau$ differs across samples and action types.
    \emph{Right}: Providing more history does not necessarily yield the optimal result, suggesting effective usage of historical information is under exploration.}
    \label{fig:history_length_analysis}
    \vspace{-5mm}
\end{figure*}

\section{Preliminaries}
\textbf{Problem Definition.} The goal of a GUI agent is to complete high-level tasks by interacting with graphical user interfaces through a sequence of low-level actions.  At each timestep $t$, the agent observes a natural language task instruction $I$, a current screenshot observation $s_t$, and a history context $H_t = {(s_{t-\tau}, a_{t-\tau}), \dots, (s_{t-1}, a_{t-1})}$, where $\tau$ denotes the history window size, which is the number of previous steps included in the history. The agent then generates an action $a_t$ conditioned on $(I, H_t, s_t)$ and executes it in the GUI environment. This interaction process defines a sequential decision-making problem, which can be formulated as a Markov Decision Process (MDP).
\vspace{3mm}

\textbf{Application of GRPO in GUI Agents.} When applying GRPO to training GUI agents, we treat the agent as a policy model and reorganize the inputs by defining $q = (I, H_t, s_t)$. For each input $q$, the policy model generates a set of candidate responses $O = {o_1, \dots, o_G}$, where each response $o_i$ consists of a thought process $t_i$ and a corresponding action $a_i$, i.e., $o_i = (t_i, a_i)$. The standard GRPO algorithm adopts the following training objective:
{\small
\begin{equation}
\begin{aligned}
\mathcal{L}_{GRPO} = -\mathbb{E}\Bigg[
\sum_{i=1}^{G} \Big(
    &\min\Big(
        \frac{\pi_{\theta}(o_i \mid q)}{\pi_{\theta_{\text{old}}}(o_i \mid q)}A_i,\ 
        \text{clip}\Big(
            \frac{\pi_{\theta}(o_i \mid q)}{\pi_{\theta_{\text{old}}}(o_i \mid q)},\\
            1-\epsilon, 1+\epsilon
        \Big)A_i
    \Big) 
    &- \beta\,\text{KL}
    \big(\pi_{\theta}(o_i \mid q)\,\|\,\pi_{\theta_{\text{ref}}}(o_i \mid q)\big)
\Big)
\Bigg]
\label{eq:grpo_main}
\end{aligned}
\end{equation}}
where $\epsilon$ is the clipping parameter, $\beta$ is a hyperparameter. $A_i$ is the advantage calculated by normalizing the group level rewards $\{R_i\}_{i=1}^G$. By normalization, advantage $A_i$ represents the relative quality of the $i$-th response.
\vspace{-2mm}



\section{Rethinking History Usage: Limitations of Fixed Context and the Anchoring Role of Actions}
\label{subsec:anchors}
A key challenge in building strong GUI agents is how to appropriately leverage historical context. To address this, we conduct two empirical studies focusing on effectiveness and efficiency. The first study analyzes how different samples benefit from different history lengths, showing that a fixed-length context is often suboptimal. This highlights the need for dynamically adjusting context length to match step-specific dependencies. The second study examines how historical information flows through the model and finds that action tokens serve as anchors for aggregating and delivering useful visual semantics. This highlights the importance of retaining history actions while compressing redundant visuals to improve efficiency without compromising performance.
\vspace{-3mm}

\subsection{Analyzing the Impact of History Length}
In GUI navigation, different steps may depend on varying lengths of history context. We conduct an empirical analysis to measure the impact of history length on decision quality across different samples and action types. As shown in the \textbf{\textit{left}} plot of Figure~\ref{fig:history_length_analysis}, we perform rollout-based evaluation over the training set using a base model with fixed policy weights. For each sample, we conduct $8$ rollouts under different history lengths $\tau \in \{0, 1, 2\}$, and record the average reward under each setting. We then determine the optimal history length per sample by comparing these mean rewards. Samples with negligible differences (mean reward difference $< 0.05$) are discarded. The result reveals a clear pattern: different samples exhibit different optimal history lengths. While some steps benefit from short-term context, others require longer to achieve higher reward. 

Notably, we further analyze cases where shorter history lengths outperform longer ones, as shown in the \textbf{\textit{right}} plot of Figure~\ref{fig:history_length_analysis}. For each such sample, we compute the mean-reward difference between the better-performing shorter context and a longer one, i.e., $\mathrm{Improvement} = \mathrm{mean\_reward}(\tau_{\text{short}}) - \mathrm{mean\_reward}(\tau_{\text{long}})$. The resulting distribution illustrates that these improvements are non-trivial, shorter histories can yield significantly higher rewards in certain situations. This phenomenon suggests that longer history is not always more beneficial, and in some cases may even hinder performance (lower reward), likely due to introducing irrelevant information. Together with the left plot, this reinforces the insight that a fixed context length $\tau$ cannot accommodate the diverse temporal dependencies across samples. Dynamically varying the length of historical context can lead to more effective model behavior and improved decision quality.

\begin{figure*}[h]
    \centering
    \includegraphics[width=0.98\textwidth]{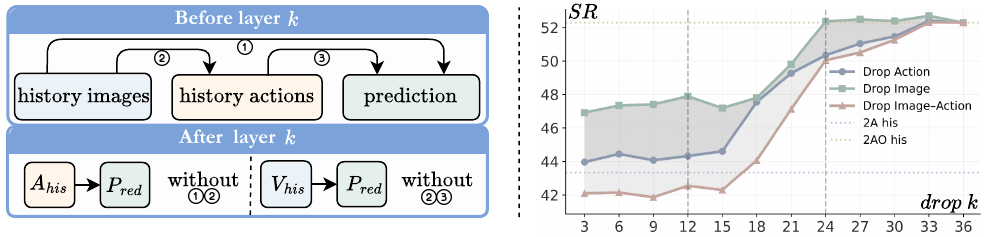}
    \vspace{-2mm}
    \caption{\textbf{Layer-wise token-drop analysis.}
    \emph{Left:} Schematic of the layer-wise token-drop probe, illustrating the information flow of image-drop and action-drop.
    \emph{Right:} Dropping $A_{\mathrm{his}}$ at shallow depths ($k{<}12$) causes a much larger decline than dropping $V_{\mathrm{his}}$. Even
    if rich visual information is retained, later layers cannot directly extract effective cues from $V_{\mathrm{his}}$ without the action anchors. As \(k\) increases, the action-drop curve rises toward the image-drop curve and the image–action drop curve converges rapidly. }
    \label{fig:info_flow}
    \vspace{-4mm}
\end{figure*}

\subsection{History Actions as Information Flow Anchors}
Naively appending historical observations to the context inflates sequence length and computation cost, burdening the model with tokens that contribute little to decision making~\citep{zhang2025progrm}. In practice, only the regions associated with past actions tend to carry meaningful semantics for decision-making, while most other areas tend to be redundant. This motivates designing a method to compress redundant historical information to improve efficiency without harming performance. The central question is which tokens should be preserved during layer interaction. Different from previous work~\citep{chen2024image, chen2025less} that adopts conclusions from information flow studies of single-image VQA scenario~\citep{zhang2025cross} to identify tokens for retention, we conduct an information flow analysis in the GUI navigation scenario within a reinforcement learning framework, tracing how historical visuals and actions interact across model depth.

\textbf{Layer-wise token-drop setup.}
We probe how history propagates through the LLM with a layer-conditioned token drop. Qwen2.5-VL-3B (36 layers) is trained on our dataset (Section ~\ref{experiments}) with history image/action context and evaluated on AndroidControl using step successful rate (SR). At depth $k$, we remove from layer $k{+}1$ onward: (i) history actions $A_{\mathrm{his}}=\{a_{t-\tau},\ldots,a_{t-1}\}$; (ii) history images $V_{\mathrm{his}}=\{s_{t-\tau},\ldots,s_{t-1}\}$; or (iii) both history actions and images $H_t= {(s_{t-\tau}, a_{t-\tau}), \dots, (s_{t-1}, a_{t-1})}$. Sweeping $k$ yields the curves in Figure~\ref{fig:info_flow}; gaps between them show that later layers access history chiefly via action tokens, with a smaller direct contribution from visuals.

%
\textbf{Shallow depth ($k{<}12$):}
As shown in Figure~\ref{fig:info_flow}, dropping $A_{\mathrm{his}}$ in shallow layers causes a large performance degradation, while dropping $V_{\mathrm{his}}$ at the same depths is much less harmful. 
This pattern indicates that effective use of historical information \emph{depend on the action tokens as the anchor}: even if rich visual information is retained, later layers cannot directly extract effective cues from $V_{\mathrm{his}}$ without the action anchors. 
By contrast, dropping $V_{\mathrm{his}}$ in this range is more tolerable, since the model has already merged part of the visual history into $A_{\mathrm{his}}$; the action anchors then pass that information forward through subsequent layers.

\textbf{Deeper depth ($k{\ge}12$):}
For $k>12$, the \emph{action-drop} curve rises quickly and, by mid depth, its gap to the \emph{image-drop} curve is small. 
This indicates that, in these layers, predictions obtain historical information through interactions with the action anchors. 
Once that interaction has already occurred before the drop point, removing $A_{\mathrm{his}}$ or $V_{\mathrm{his}}$ later causes only little decline, and performance continues to improve with depth. When \(k \ge 24\), all three curves converge to the no-compression accuracy, indicating that the model has already integrated most historical information and can proceed without retaining these history tokens.

\textbf{Key discovery.}
From the layer-wise drop analysis in Figure~\ref{fig:info_flow}, relying only on history actions (2A) yields limited benefit: without their paired screenshots, history action tokens \(A_{\mathrm{his}}\) lack grounded semantics and provide little guidance for decision-making. In contrast, incorporating history observations (2AO) proves critical for navigation success. However, the gain does not arise from directly attending to history images. Instead, it emerges mainly at intermediate depths, where \(A_{\mathrm{his}}\) interacts with the history visual tokens \(V_{\mathrm{his}}\) and delivers the extracted information to subsequent tokens, highlighting the role of history actions as anchors for multimodal information flow. 

\textbf{Implications for designing compression methods.}
Guided by the above experimental results and analysis, we adopt a single rule to balance efficiency and effectiveness: \emph{compress history only after an early fusion depth $k$}, pruning $V_{\mathrm{his}}$ while \emph{retaining} $A_{\mathrm{his}}$. 
On the \textit{efficiency} side, removing history images greatly reduces sequence length and computation cost; on the \textit{effectiveness side}, keeping the action anchors preserves the historical cues that later tokens actually use. To make the proposed compression method work better, training is essential to strengthen cross-modal interaction in the first $k$ layers, so that by the time compression occurs, the relevant visual context has been sufficiently transferred into $A_{\mathrm{his}}$, thereby preserving the necessary information flow.
\begin{figure*}[t]
    \centering
    \setlength{\abovecaptionskip}{0mm}
    \setlength{\belowcaptionskip}{0mm}
    \includegraphics[width=1.0\textwidth]{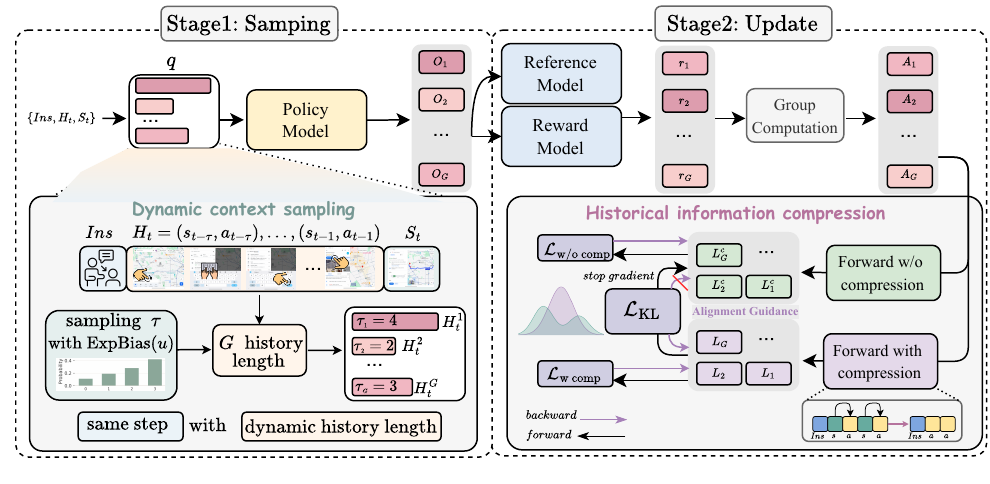}
\caption{Overview of our history context-aware optimization framework for building HiconAgent. HCPO improves both the sampling and update phases of policy optimization by incorporating two key components: (1) \textbf{Dynamic Context Sampling (DCS)}, which introduces varied history lengths during training to encourage context-effective decision-making, and (2) \textbf{Anchor-guided History Compression (AHC)}, which adopts a dual-branch architecture where both branches share sampled responses and group-wise advantages. The compressed branch is trained using policy gradients, aligned with the uncompressed branch via a history-enhanced alignment loss.}
\label{fig:framework}
\vspace{-3mm}
\end{figure*}

\section{HiconAgent}
We propose History Context-aware Policy Optimization (HCPO), a reinforcement learning framework that improves history utilization by strengthening both the sampling and update phases. Dynamic Context Sampling (DCS) varies history lengths during rollouts to guide effective context usage, while Anchor-guided History Compression (AHC) reduces redundancy during updates via alignment with full-history supervision. Together, they enable policy optimization with effective and efficient historical context usage.

\subsection{Dynamic Context Sampling}
As shown in the analysis in Section~\ref{subsec:anchors}, relying on a fixed-length history context is often suboptimal. This motivates adapting the history length to step-specific dependencies, allowing the policy to leverage the appropriate amount of context as needed. To address this, we propose Dynamic Context Sampling, which exposes the policy to diverse history lengths during training. This allows the agent to adaptively focus on the most relevant context, thereby improving overall policy learning.

During training, instead of always feeding a fixed-length history $H_t$, we dynamically sample and select $G$ variants of truncated histories $\{H_t^{1}, \dots, H_t^{G}\}$, where $H_t^{i}$ uses a selected history length $\tau_i \leq \tau$ sampled from distribution $p$. However, naively adopting a uniform distribution leads to a degeneration phenomenon, as later shown in our sampling ablation. Motivated by this empirical observation, we propose an exponential-biased distribution that mitigates training collapse by gradually shifting the sampling toward larger $\tau$ as training progresses. At training step $u$, our exponential-biased distribution $\text{ExpBias}(u)$ is defined as:
\begin{equation}
P(\tau_i \mid u) = \frac{\exp\big(\lambda(u)\, \tau_i\big)}{\sum_{j=0}^{N} \exp\big(\lambda(u)\, j\big)}
\label{eq:dcs_prob_main}
\end{equation}
where $\lambda(u)$ is a linear function that increases with $u$. In the early stage of training, $\lambda(u)\approx 0$ and the distribution is nearly uniform, encouraging random exploration. As training progresses, $\lambda(u)$ gradually grows, yielding an increasingly biased exponential distribution that favors larger values of $\tau_i$. This schedule smoothly shifts the sampling strategy from random selection to full-context history. 

Each variant forms an input $q_i = (I, H_t^{i}, s_t)$, and produces a response $o_i$. These $G$ responses are evaluated as a group, yielding an advantage value for each response. Responses with higher advantages receive stronger gradient updates, allowing the policy to adaptively learn which history lengths lead to improved policy behavior. Importantly, to maintain consistency between training and inference, we combine each sampled response $o_i$ with the full history context input $(I, H_t, s_t)$ to compute the logits for optimization. This design allows the policy to explore effective solutions under varying historical conditions while being evaluated under a unified context length.
%

\subsection{Anchor-guided History Compression}
Leveraging the observation in Section ~\ref{subsec:anchors} that history actions preserve the historical cues that later tokens actually use, we propose Anchor-Guided History Compression. AHC keeps past actions as anchors, prunes visual history after early fusion, and leverages alignment with an uncompressed branch, preserving decision quality while reducing sequence length and FLOPs.

Let \( q = \{I, H_t, s_t\} \) be the model input at step \(t\), and define the importance sampling ratio as $\rho_i = \frac{\pi_{\theta}(o_i \mid q)}{\pi_{\theta_{\text{old}}}(o_i \mid q)}$. The responses from the uncompressed branch are optimized using the standard GRPO objective:
{\small
\begin{equation}
\begin{split}
\mathcal{L}_{\text{w/o comp}}
= -\sum_{i=1}^G \Big(
&\min\big( \rho_i A_i,\ \text{clip}(\rho_i, 1-\epsilon, 1+\epsilon)\,A_i \big) \\
&- \beta\, \text{KL}\big[
\pi_{\theta}(o_i \mid q)\,\|\,\pi_{\theta_{\text{ref}}}(o_i \mid q)
\big]
\Big),
\label{eq:grpo_loss}
\end{split}
\end{equation}
}
To reduce redundancy, we remove all history vision tokens $V_{\mathrm{his}}=\{s_{t-\tau}, \dots, s_{t-1}\}$ while retaining the past action tokens. The model continues forwarding with the compressed sequence $\{I, A_{\mathrm{his}}, s_t\}$, 
where the retained action tokens $A_{\mathrm{his}}=\{a_{t-\tau}, \dots, a_{t-1}\}$ constitute the compressed history $H_t^c$. The compressed branch is further optimized with a GRPO-style objective using the same responses and advantages, leveraging the superior quality of responses generated from the uncompressed branch, where $q^c = \{I, H^c_t, s_t\}$ denotes the inputs with compressed history and the importance sampling ratio as $\rho_i^c = \frac{\pi_{\theta}(o_i \mid q^c)}{\pi_{\theta_{\text{old}}}(o_i \mid q^c)}$.
{\small
\begin{equation}
\begin{split}
\mathcal{L}_{\text{w/ comp}} = -\sum_{i=1}^G \bigg(
    \min\left( \rho^c_i A_i,
    \text{clip}(\rho^c_i, 1-\epsilon, 1+\epsilon) A_i \right) \\
    -\; \beta\, \text{KL}\big[
    \pi_{\theta}(o_i \mid q^c)\,\|\,
    \pi_{\theta_{\text{ref}}}(o_i \mid q^c)
    \big]
\bigg).
\end{split}
\end{equation}}
To ensure that the compressed branch retains the core decision-making ability of the original model, we introduce a \textit{history-enhanced alignment objective}. Specifically, given the original history \( H_t \) and its compressed counterpart \( H^c_t \), we perform parallel forward passes through both branches. We then minimize the KL divergence between their output distributions, effectively using the uncompressed branch as a teacher to guide the compressed branch. Note that the uncompressed branch is used only for guidance; we detach its outputs to prevent gradient backpropagation in the KL loss. This approach allows the compression module to reduce redundancy while preserving critical behavioral patterns from the original branch. The alignment objective is defined as:
\begin{equation}
\mathcal{L}_{\text{KL}} = \sum_{i=1}^G \text{KL} \left[ 
\pi_{\theta}(o_i \mid q^c) \,\|\, \pi_{\theta}(o_i \mid q)  
\right],
\end{equation}
The final HCPO loss is the sum of the uncompressed branch loss, the compressed branch loss, and the alignment constraint:
\begin{equation}
\mathcal{L}_{\text{HCPO}} = \mathcal{L}_{\text{w/o comp}} + \mathcal{L}_{\text{w comp}} + \lambda \mathcal{L}_{\text{KL}},
\end{equation}
where $\lambda$ controls the strength of the alignment guidance. This framework enables effective policy optimization under compressed context while preserving temporal consistency.

        
        
        
        
        


\subsection{Reward Design}
In GUI navigation tasks, each action consists of a type and a value. The type is chosen from a set of discrete options (e.g., CLICK, SCROLL), while the value format varies depending on the type. Based on this characteristic of GUI tasks, we introduce the following three rewards:

\textbf{Format reward ($r^f$):} This term ensures the model’s response adheres to a predefined structure like \texttt{<think>...</think><answer>...</answer>.} It returns 1 if the format is correct, 0 otherwise, promoting structured reasoning and output generation.

\textbf{Action type reward ($r^t$):} We assign 1 if the predicted action type exactly matches the ground-truth type, and 0 otherwise. This term enforces correctness at the semantic level of action selection.

\textbf{Action value reward ($r^v$):} For actions \textit{without} values (e.g., PRESS\_BACK), the reward is 1 if the type is correct. For actions with \textit{textual} values (e.g., TYPE, OPEN\_APP), we compute the F1 score between prediction and ground truth, awarding 1 if F1 $>$ 0.5. For actions with \textit{discrete} values (e.g., SCROLL), the value must exactly match the ground truth. For \textit{coordinate-based} actions (e.g., CLICK), we calculate the Euclidean distance $d$ between predicted and actual coordinates, and assign a continuous reward $r^v = 1 - d$ to allow fine-grained feedback for grounding accuracy.

The final reward used for policy optimization is the sum of the three components:
\begin{equation}
r = r^f + r^t + r^v
\end{equation}



\section{Experiments}
\label{experiments}
\subsection{Implementation Details}
\textbf{Metrics.} We evaluate our model on three representative navigation-oriented datasets, AndroidControl-High\cite{li2024effects}, AITW\cite{lai_autowebglm_2024} and GUI-Odyssey\cite{lu_gui_2024}, considering only their test splits under an out-of-distribution (OOD) evaluation setting to assess generalization performance. We use three standard metrics widely adopted in prior work on GUI agents, following the evaluation protocol of Os-Atlas~\citep{wu_os-atlas_2024}. Specifically, we report action type prediction accuracy (\textbf{Type}), GUI grounding accuracy (\textbf{Grounding}), and step success rate (\textbf{SR}). 

FLOPs is computed using deepspeed flops-profiler with batch size 1. We include all model components when calculating FLOPs. The reported number is averaged over 200 samples from the training set with history length $\tau=2$.

\textbf{Training and Evaluation.} HiconAgent-3B is built upon Qwen2.5-VL-3B. We follow the same training setting in GUI-R1~\citep{luo2025gui}. In the reinforcement fine-tuning stage, we adopt the following hyperparameter settings to ensure stable optimization. The rollout batch size and global batch size is set to 64, with 8 rollouts per update step. We use a small learning rate of $1 \times 10^{-6}$ to stabilize training. To balance computational efficiency and generation quality, the maximum number of input pixels is capped at 1,003,520. For the language input and output, both the maximum prompt length and the maximum response length are restricted to 2048 tokens. The rollout temperature is fixed at 1.0 to encourage diverse yet consistent exploration during training.

To keep the number of tokens consistent with prior work, we set the history window size to 2, meaning the agent can access up to two past interaction steps. Each historical step includes both the screenshot observation and the corresponding action, i.e., \(H_t = \{(s_{t-2}, a_{t-2}), (s_{t-1}, a_{t-1})\}\). All images are resized to a fixed resolution, and the number of visual tokens after encoding is limited to a maximum of 512 to ensure computational efficiency.



\textbf{Model Configuration.} From an efficiency perspective, HiconAgent-3B adopts the drop $k=6$ configuration by default in the following experiments.
This setting achieves up to a 60\% reduction in FLOPs while maintaining competitive accuracy, as shown in Table~\ref{tab:drop_ablation_combined}. 

\subsection{Experiment Results}
\definecolor{improve}{RGB}{120,180,130} 
\definecolor{drop}{RGB}{200,90,90}      
\definecolor{rowgray}{RGB}{225,213,231}

\definecolor{drop}{RGB}{120,180,130} 
\definecolor{improve}{RGB}{200,90,90} 
\definecolor{rowgray}{RGB}{225,213,231}
We present the main experimental results in Table~\ref{tab:main_results} and Table~\ref{tab:gui_half_table} on three representative GUI navigation datasets: AndroidControl-High~\citep{li2024effects}, AITW~\citep{rawles2023androidinthewild} and GUI-Odyssey~\citep{lu_gui_2024}. Table~\ref{tab:main_results} provides a detailed comparison under the same data scale and training settings, highlighting the effect of our history-aware optimization strategy against both supervised fine-tuning and reinforcement fine-tuning baselines. Table~\ref{tab:gui_half_table} further extends the comparison to recent advanced GUI agents of varying model sizes and training data volumes, demonstrating the generalization ability of our approach in out-of-distribution (OOD) scenarios.

\begin{table}[t]
\centering
\small
\setlength{\tabcolsep}{2pt}
\renewcommand{\arraystretch}{1.05}
\resizebox{\linewidth}{!}{%
\begin{tabular}{c|cc|cc}
\toprule[1pt]
\multirow{2}{*}{\textbf{Models}} & \multicolumn{2}{c}{\textbf{AC-High}} & \multicolumn{2}{c}{\textbf{GUI-Odyssey}} \\
\cmidrule(lr){2-5}
 & Grounding & SR & Grounding & SR \\
\midrule
\rowcolor[HTML]{F5F5F5}
\multicolumn{5}{c}{\textbf{Supervised Fine-Tuning}} \\ 
\midrule
GUI-R1-3B \cite{luo2025gui}      & 49.53 & 41.22 & 32.21 & 27.31 \\
GUI-R1-7B \cite{luo2025gui}      & 58.69 & 48.11 & 38.65 & 34.44 \\
Qwen2.5VL-3B                     & 52.89 & 41.95 & 36.21 & 31.74 \\
\midrule
\rowcolor[HTML]{F5F5F5}
\multicolumn{5}{c}{\textbf{Zero-Shot}} \\ 
\midrule
GPT-4o \cite{hurst2024gpt}       & 30.90 & 21.17 & 14.17 &  5.36 \\
Qwen2.5VL-3B                     & 46.51 & 38.90 & 26.49 & 26.69 \\
Qwen2.5VL-7B                     & 59.71 & 47.06 & 37.78 & 34.37 \\
\midrule
\rowcolor[HTML]{F5F5F5}
\multicolumn{5}{c}{\textbf{Reinforcement Fine-Tuning}} \\ 
\midrule
GUI-R1-3B \cite{luo2025gui}      & 56.24 & 46.55 & 41.52 & 41.33 \\
GUI-R1-7B \cite{luo2025gui}      & \textbf{65.56} & 51.67 & 43.64 & 38.79 \\
\rowcolor[HTML]{E9E4F0}
\textbf{HiconAgent-3B}          & 65.51 {\tiny\color{drop}(-0.05)} & 
\textbf{52.40} {\tiny\color{improve}(+0.73)} & 
\textbf{52.10} {\tiny\color{improve}(+8.46)} & 
\textbf{50.11} {\tiny\color{improve}(+11.32)} \\
\bottomrule[1pt]
\end{tabular}}
\vspace{-2mm}
\caption{Performances on \textbf{AndroidControl-High} and \textbf{GUI-Odyssey}.
\textcolor{improve}{Red} indicates improvement, 
\textcolor{drop}{green} indicates degradation compared to GUI-R1-7B. Our 3B model outperforms GUI-R1-7B by \textbf{+8.46}\% grounding and \textbf{+11.32}\% SR on GUI-Odyssey.}
\vspace{-5mm}
\label{tab:main_results}
\end{table}


As shown in Table~\ref{tab:main_results}, HiconAgent-3B achieves consistent improvements over all baselines trained under the same data scale and training settings. We first observe that HiconAgent-3B consistently outperforms the supervised baseline Qwen2.5VL-3B, with clear gains in both grounding accuracy and step successful rate. When trained with reinforcement learning, HiconAgent-3B further surpasses the GRPO baseline GUI-R1-3B (+5.85\% SR) and GUI-R1-7B (+0.73\% SR) on AndroidControl under the same training setup, demonstrating a stronger decision-making ability.

The advantage of HiconAgent becomes especially pronounced on the long-horizon GUI-Odyssey benchmark. 
Despite having less than half the parameters of GUI-R1-7B, our 3B model achieves a remarkable +8.46\% improvement in grounding accuracy and +11.32\% in step successful rate. 
Compared with GUI-R1, which does not explicitly exploit historical information, HiconAgent-3B adopts a more effective strategy for leveraging historical context, leading to stronger sequential reasoning and consistent execution.

\definecolor{oodgreen}{RGB}{220, 255, 220}
\definecolor{idred}{RGB}{255, 230, 230}
\renewcommand{\arraystretch}{1.05}
\begin{table}[b]
\vspace{-3mm}
\centering
\small
\setlength{\tabcolsep}{0.8pt}
\begin{tabular}{l|cc|ccc|c}
\toprule
\textbf{Model} & \textbf{\#Data}& \textbf{Filter}& \textbf{AC-High} & \textbf{AITW} & \textbf{Odyssey} & \textbf{Avg SR} \\
\midrule
GPT-4o \cite{hurst2024gpt} & -- & -- &21.17 & 26.07 & 5.36 & 17.54 \\
OS-Atlas-7B \cite{wu_os-atlas_2024}& 13M &\checkmark &\cellcolor{oodgreen}29.83 & \cellcolor{oodgreen}41.38 &\cellcolor{oodgreen}26.96 & 32.72 \\
GUI-R1-7B \cite{luo2025gui}& 3K &\checkmark &\cellcolor{idred}51.67 & \cellcolor{oodgreen}\textbf{55.31} & \cellcolor{oodgreen}38.79 & 48.59 \\
UI-shift-7B \cite{gao2025uishift}& 2K &\checkmark &\cellcolor{idred}52.16 & \cellcolor{oodgreen}54.38 & \cellcolor{oodgreen}32.75 & 46.43 \\
UI-AGILE \cite{lian2025ui}& 9K &\checkmark &\cellcolor{idred}50.97 & \cellcolor{idred}48.66 & \cellcolor{idred}36.96 & 45.53 \\
infiGUI-3B \cite{liu2025infigui}& 32K &\checkmark &\cellcolor{idred}\textbf{71.10} & \cellcolor{oodgreen}46.51 & \cellcolor{oodgreen}33.15 & 50.25 \\
\midrule
\textbf{HiconAgent-3B} & 3K & \ding{55} &\cellcolor{oodgreen}52.40 & \cellcolor{oodgreen}51.91 & \cellcolor{oodgreen}\textbf{50.11} & \textbf{51.47} \\
\bottomrule
\end{tabular}
\caption{Step Successful Rate evaluated on three representative GUI navigation datasets, compared with recent models. Cells highlighted in red correspond to datasets that are IID for the respective models, whereas cells highlighted in green indicate OOD setting.}
\label{tab:gui_half_table}
\end{table}

Table~\ref{tab:gui_half_table} summarizes OOD generalization across the three benchmarks. Without relying on large-scale data filtering or dataset curation, HiconAgent is trained on only 3K unfiltered samples, yet it achieves the highest average step successful rate (51.47\%) among all compared models. Remarkably, it surpasses stronger models such as OS-Atlas-7B, GUI-R1-7B, and infiGUI-3B, which are trained with much larger data volumes (13M-filtered, 3K-filtered, and 32K-filtered respectively). Although infiGUI-3B achieves a notably high SR on the AndroidControl benchmark (71.1\%), its performance drops considerably on AITW and GUI-Odyssey, indicating weaker generalization under OOD conditions. This demonstrates that our HCPO framework is not only effective but also highly data-efficient, enabling strong generalization even without large-scale data curation.


\begin{table}[t]
    \centering
    \footnotesize
    \setlength{\tabcolsep}{1.2pt}
    \renewcommand{\arraystretch}{1.25}
    \begin{tabular}{l|ccccc}
    \toprule[1pt]
    Models & DCS & Update $\tau$ & Sampling $p(\tau)$ & AC-High SR & Train Hrs \\
    \midrule
    HCPO (w/o DCS) & -- & 2 & -- & 51.03 & 17h \\
    HCPO (Uniform) & \checkmark & 2 & $U(0,2)$ & 50.53 & 17h \\
    HCPO (Uniform) & \checkmark & \{0,1,2\} & $U(0,2)$ & 51.62 & 30h \\
    \rowcolor{rowgray}
    HCPO (ExpBias) & \checkmark & 2 & $\text{ExpBias}(u)$ & 52.40 & 17h \\
    \bottomrule[1pt]
    \end{tabular}
    \caption{Ablation study on different sampling distributions in Dynamic Context Sampling (DCS). 
    $U(0,2)$ denotes uniform distribution over $\tau\in \{0,1,2\}$, while $\text{ExpBias}(u)$ represents the exponential-biased schedule defined in Eq.~(\ref{eq:dcs_prob_main}).}
    \label{tab:dcs_sampling_ablation}
    \vspace{-6mm}
\end{table}

\subsection{Ablation Study}
\definecolor{drop}{RGB}{120,180,130} 
\definecolor{improve}{RGB}{200,90,90}      
\definecolor{rowgray}{RGB}{225,213,231}

\begin{figure}[H]
    \vspace{-2mm}
    \centering
    \includegraphics[width=0.47\textwidth]{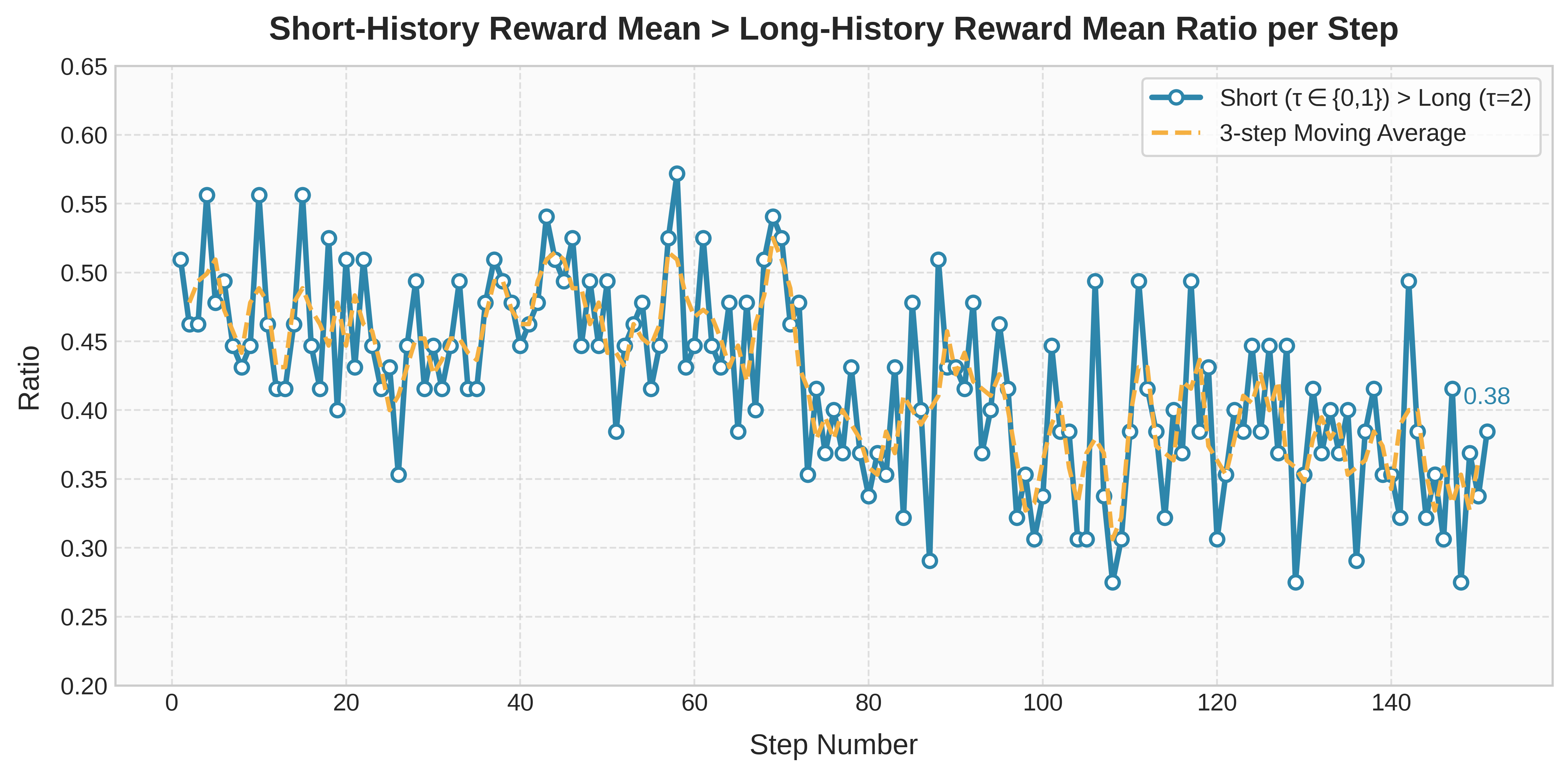}
    \vspace{-3mm}
\caption{Evolution of the short-vs-long history reward ratio under uniform $\tau$ sampling. 
The declining ratio reflects the gradual degradation of short-history response quality during training.}
\label{fig:uniform_evolve}
\vspace{-4mm}
\end{figure}

\textbf{Impact of sampling distribution $p$.} We perform an ablation on the choice of $p(\tau)$ to study how different sampling strategies affect learning. A uniform sampling baseline $U(0,2)$ is first adopted for comparison, where $\tau \!\in\!\{0,1,2\}$ denotes the sampled history length. Under plain uniform sampling $U(0,2)$, we observe a degeneration phenomenon during training: the sampling quality of the shorter histories ($\tau=0,1$) weakens over time as shown in Figure ~\ref{fig:uniform_evolve}. This is because we only compute gradients and update parameters using the context with $\tau=2$ to ensure training–inference consistency. To mitigate this, we design a variant that enforces the inclusion of all input contexts with $\tau \in \{0,1,2\}$ in the optimization. While this all-$\tau$ update strategy partially recovers the performance lost by naive uniform sampling, it substantially increases training overhead and still fails to match the performance of our proposed exponential-biased distribution. The exponential-biased sampling $\text{ExpBias}(u)$ mitigates collapse by gradually biasing the sampling toward larger $\tau$ as training progresses, while still encouraging exploration of shorter histories in the early stage. This strategy achieves the best trade-off between performance and computational cost, as reported in Table~\ref{tab:dcs_sampling_ablation}.

\paragraph{Effect of DCS and AHC.}
Table~\ref{tab:branch_kl_ablation} evaluates the impact of Dynamic Context Sampling (DCS) and Anchor-guided History Compression (AHC) across the three navigation benchmarks. Training only a compressed branch using standard GRPO yields the weakest performance, indicating that unguided compression fails to retain useful historical signals. Adding an uncompressed branch without KL improves performance, showing that full-history responses help guide the compressed branch. Incorporating KL alignment between the two branches further enhances performance, validating the benefit of matching compressed outputs to the full-history teacher. Finally, enabling DCS achieves the best overall results, confirming that DCS enhances the model’s ability to utilize historical context more effectively.

\begin{table}[h]
    \centering
    \footnotesize
    \setlength{\tabcolsep}{0.7pt}
    \renewcommand{\arraystretch}{1.3}
    \begin{tabular}{l|ccc|ccc}
    \toprule[1pt]
    Models & 2Branch. & KL & DCS & AC-High & AITW & Odyssey \\
    \midrule
    GRPO & -- & -- & -- & 44.89 & 45.62 & 43.21 \\
    HCPO (w/o KL, DCS) & \checkmark & -- & -- & 
        48.70{\tiny\textcolor{improve}{(+3.81)}} & 
        49.23{\tiny\textcolor{improve}{(+3.61)}} & 
        47.09{\tiny\textcolor{improve}{(+3.88)}} \\
    HCPO (w/o DCS) & \checkmark & \checkmark & -- & 
        51.03{\tiny\textcolor{improve}{(+6.14)}} & 
        50.78{\tiny\textcolor{improve}{(+5.16)}} & 
        48.68{\tiny\textcolor{improve}{(+5.47)}} \\
    \rowcolor{rowgray}
    HCPO & \checkmark & \checkmark & \checkmark & 
        52.40{\tiny\textcolor{improve}{(+7.51)}} & 
        51.91{\tiny\textcolor{improve}{(+6.29)}} & 
        50.11{\tiny\textcolor{improve}{(+6.90)}} \\
    \bottomrule[1pt]
    \end{tabular}
    \caption{Ablation study on the dual-branch architecture, alignment loss, and DCS, evaluated using the SR metric. Experiments are conducted with compression enabled.}
    \label{tab:branch_kl_ablation}
    \vspace{-2mm}
\end{table}

\begin{figure*}[h]
    \centering
    \includegraphics[width=1\textwidth]{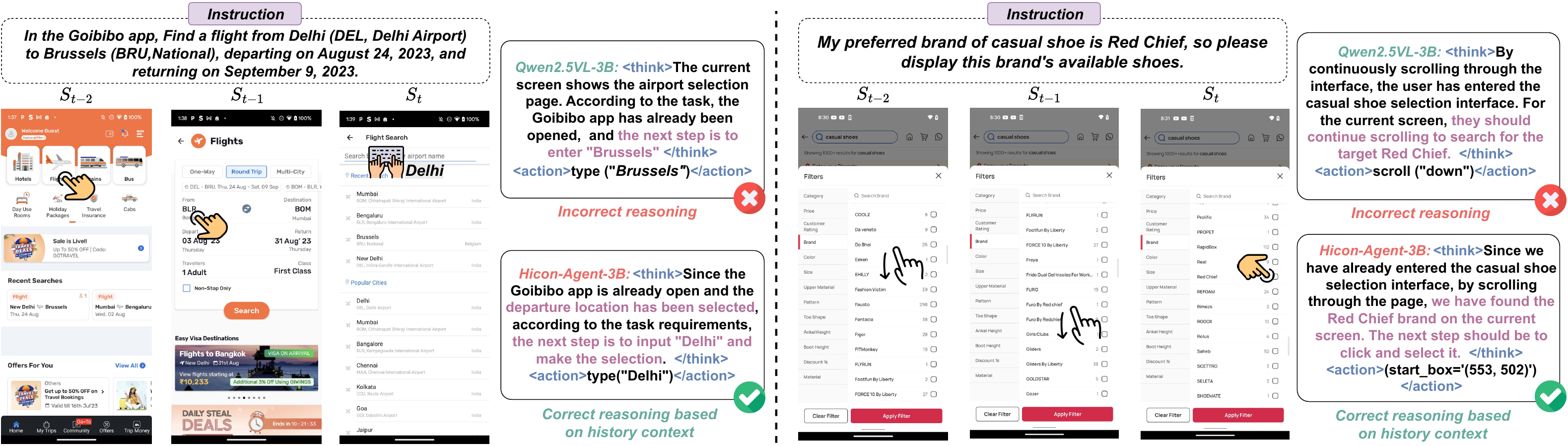}
\caption{To illustrate HCPO’s enhancement in leveraging historical information, we present two scenarios:
Left (Flight Booking) and Right (Shopping Task). Our model correctly inputs \textit{Delhi} by reasoning over historical context and selects \textit{Red Chief} despite visual redundancy. While the base model trained without HCPO misinterprets history and fails in both cases.
}
    \label{fig:case_study}
\vspace{-3mm}
\end{figure*}

\textbf{Impact of layer drop position.} We study how the position of the layer drop ($k$) affects the trade-off between computation and performance. As shown in Table~\ref{tab:drop_ablation_combined}, dropping earlier layers (e.g., $k=1$) leads to the largest FLOPs reduction (-33.81 \% vs 3B), but comes at the cost of degraded performance across all metrics. As $k$ increases, computation gradually increases while performance steadily improves. Our choice of $k=6$ offers a good balance, retaining competitive performance with a 59.54\% FLOPs reduction and 2.47× computational speedup (62.31T / 35.75).
\definecolor{softchg}{HTML}{8A5FA2} 
\begin{table}[h]
  \centering
  \footnotesize
  \setlength{\tabcolsep}{5pt}
  \renewcommand{\arraystretch}{1.1}
  \begin{tabular}{lcccccc}
    \toprule
    & \multicolumn{4}{c}{Computation} & \multicolumn{2}{c}{Performance} \\
    \cmidrule(lr){2-5} \cmidrule(lr){6-7}
    Drop $k$ & {FLOPs} & {$\Delta$ vs 3B} & {$\Delta$ vs 7B} & Tokens & Type & SR \\
    \midrule
    $k=1$    & 23.66 & {\color{softchg}-33.81\%} & {\color{softchg}-62.02\%} & 674 & 65.33 & 46.11 \\
    $k=3$    & 24.28 & {\color{softchg}-32.07\%} & {\color{softchg}-61.04\%} & 674 & 66.00 & 46.92 \\
    $k=6$    & 25.21 & {\color{softchg}-29.47\%} & {\color{softchg}-59.54\%} & 674 & 66.56 & 47.34 \\
    $k=12$   & 27.07 & {\color{softchg}-24.28\%} & {\color{softchg}-56.57\%} & 674 & 67.02 & 47.89 \\
    \rowcolor[gray]{0.9}
    w/o drop & 35.75 & - & {\color{softchg}-42.64\%} & 1664 & 69.29 & 52.29 \\
    \bottomrule
  \end{tabular}
  \caption{FLOPs, token counts, and performance of base model under different layer-drop settings, with relative FLOPs reduction versus the 3B model (35.75T) and 7B model (62.31T).}
  \label{tab:drop_ablation_combined}
  \vspace{-3mm}
\end{table}

\begin{figure}[h]
    \centering
    \includegraphics[width=0.5\linewidth]{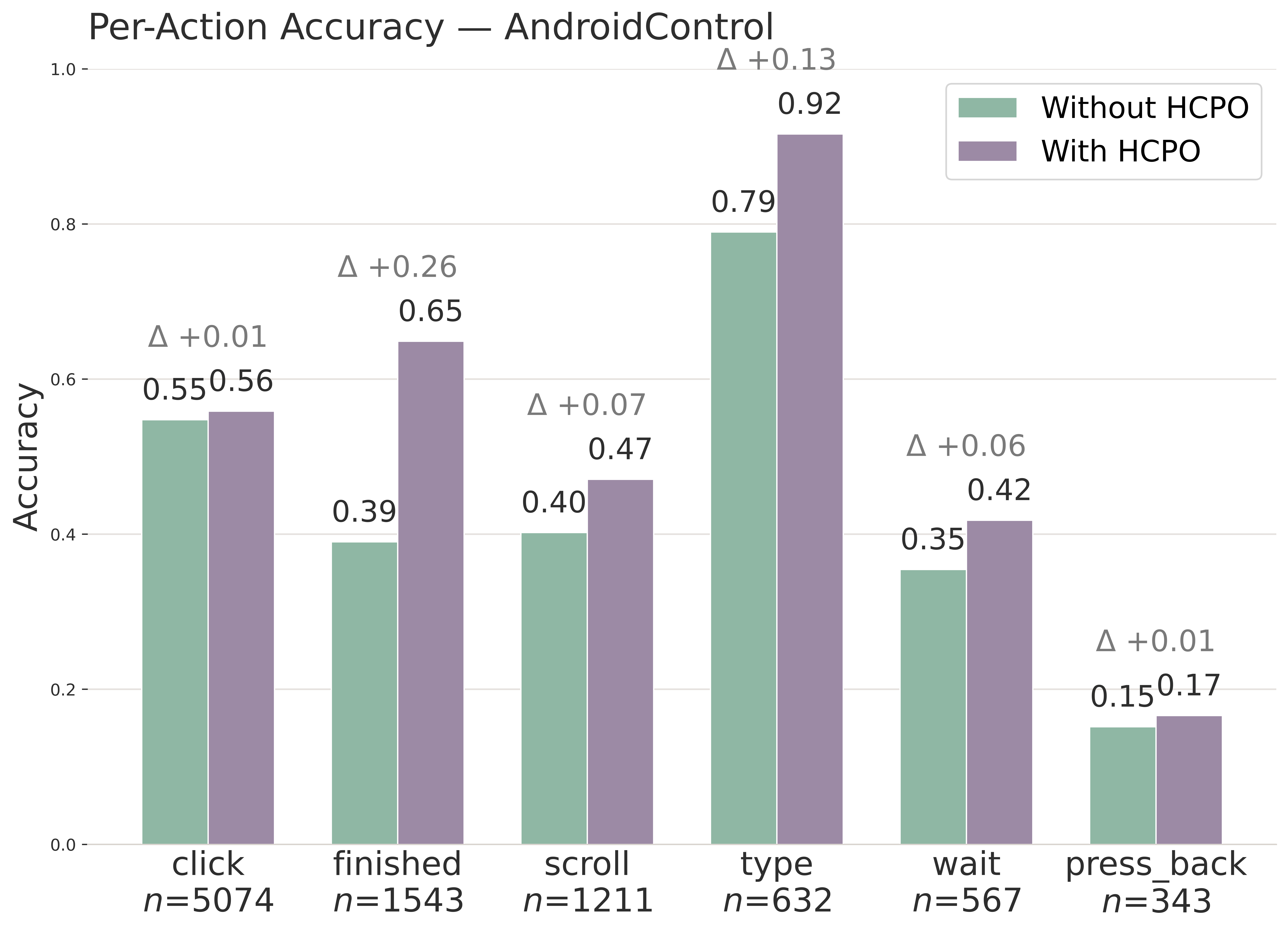}%
    \hfill
    \includegraphics[width=0.5\linewidth]{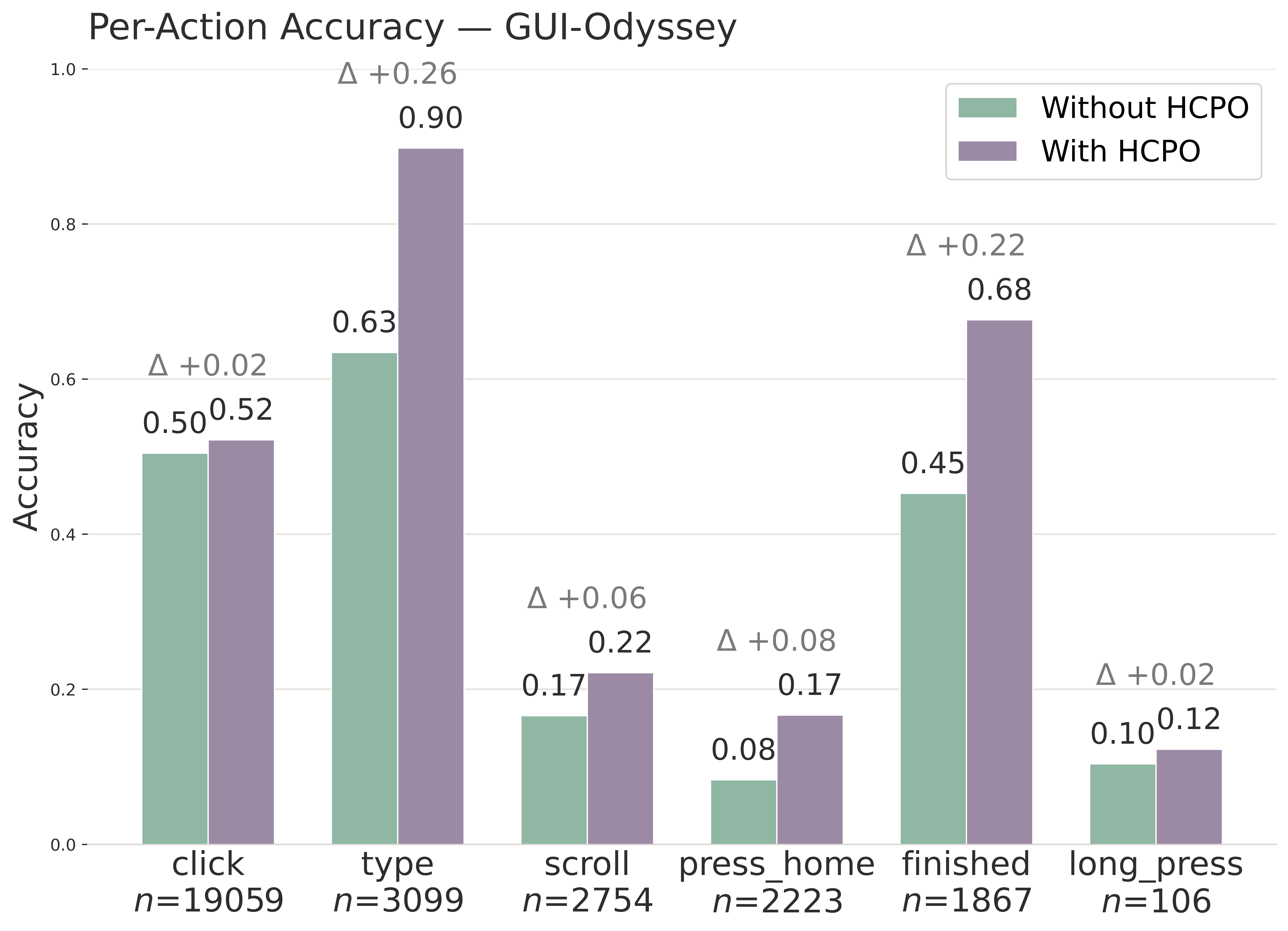}
    \caption{Per-action accuracy comparison before and after applying HCPO. Both AndroidControl (left) and GUI-Odyssey (right) benefit from history-compressed optimization, especially on \textit{finished} actions, showing improved sequential decision quality.}
    \label{fig:hcpo_ablation}
    \vspace{-3mm}
\end{figure}

\textbf{Impact of HCPO on History Utilization across Action Types}. Beyond the overall improvements reported in Table \ref{tab:branch_kl_ablation}, we further examine per-action performance to understand how HCPO influences different types of decisions. As shown in Figure~\ref{fig:hcpo_ablation}, applying HCPO consistently improves accuracy across most action categories on both AndroidControl and GUI-Odyssey. The gains are especially pronounced on finished actions, which directly determine whether a trajectory is successfully terminated, suggesting that HCPO strengthens the model’s ability to maintain global sequence control rather than overfitting to local step transitions. Improvements are also evident in challenging categories such as scroll, which demands exact direction prediction, and type, which often involves generating context-relevant input. Such findings highlight that our history-aware training paradigm remains effective in scenarios where historical cues are essential for accurate action generation. 

\textbf{Case Study on Historical Context Utilization}. 
To investigate how HCPO intuitively enhances the use of historical context, we conduct a comparative case study on two representative tasks, as shown in Figure~\ref{fig:case_study}. In the flight-booking scenario (left), HiconAgent-3B correctly reasons over previous steps and inputs \textit{Delhi} as the departure city, whereas the base model trained without HCPO mistakenly enters the destination \textit{Brussels}.  Similarly, in the shopping task (right), where repetitive history screenshots introduce visual ambiguity, our model robustly focuses on the current frame to accurately select the target brand \textit{Red Chief}, while the base model fails to resolve the ambiguity and deviates from the intended action.  These cases illustrate that HCPO enables more reliable and effective usage of historical context.
\vspace{-1mm}

\section{Conclusion}
In this paper, we present HiconAgent, a history-aware GUI agent trained with History Context-aware Policy Optimization. Through extensive empirical investigations, we first revisited how history is utilized in GUI reinforcement learning agents. Our two key studies revealed that different decision steps prefer different history lengths and historical actions serve as information flow anchors. By pairing DCS and AHC, our model outperforms larger models with fewer FLOPs. These results highlight HiconAgent as a practical path toward lightweight, high-performance GUI agents.


{
    \small
    \bibliographystyle{ieeenat_fullname}
    \bibliography{main}

@String(ICCV= {Int. Conf. Comput. Vis.})

@String(ICME = {Int. Conf. Multimedia and Expo})

@String(ICCV  = {ICCV})

@String(ICME  =	{ICME})

@article{li2025optimus,
  title={Optimus-3: Towards generalist multimodal minecraft agents with scalable task experts},
  author={Li, Zaijing and Xie, Yuquan and Shao, Rui and Chen, Gongwei and Guan, Weili and Jiang, Dongmei and Nie, Liqiang},
  journal={arXiv preprint arXiv:2506.10357},
  year={2025}
}

@inproceedings{li2025optimus2,
  title={Optimus-2: Multimodal minecraft agent with goal-observation-action conditioned policy},
  author={Li, Zaijing and Xie, Yuquan and Shao, Rui and Chen, Gongwei and Jiang, Dongmei and Nie, Liqiang},
  booktitle={Proceedings of the computer vision and pattern recognition conference},
  pages={9039--9049},
  year={2025}
}

@inproceedings{NEURIPS2021_27d52bcb,
 author = {Chen, Gongwei and Song, Xinhang and Wang, Bohan and Jiang, Shuqiang},
 booktitle = {Advances in Neural Information Processing Systems},
 editor = {M. Ranzato and A. Beygelzimer and Y. Dauphin and P.S. Liang and J. Wortman Vaughan},
 pages = {4987--4999},
 publisher = {Curran Associates, Inc.},
 title = {See More for Scene: Pairwise Consistency Learning for Scene Classification},
 url = {https://proceedings.neurips.cc/paper_files/paper/2021/file/27d52bcb3580724eb4cbe9f2718a9365-Paper.pdf},
 volume = {34},
 year = {2021}
}

@article{li2024enhancing,
  title={Enhancing emotional generation capability of large language models via emotional chain-of-thought},
  author={Li, Zaijing and Chen, Gongwei and Shao, Rui and Xie, Yuquan and Jiang, Dongmei and Nie, Liqiang},
  journal={arXiv preprint arXiv:2401.06836},
  year={2024}
}

@article{xie2025mirage,
  title={Mirage-1: Augmenting and updating gui agent with hierarchical multimodal skills},
  author={Xie, Yuquan and Li, Zaijing and Shao, Rui and Chen, Gongwei and Zhou, Kaiwen and Li, Yinchuan and Jiang, Dongmei and Nie, Liqiang},
  journal={arXiv preprint arXiv:2506.10387},
  year={2025}
}

@article{jiang2025dive,
  title={Dive: Efficient multi-view driving scenes generation based on video diffusion transformer},
  author={Jiang, Junpeng and Hong, Gangyi and Zhang, Miao and Hu, Hengtong and Zhan, Kun and Shao, Rui and Nie, Liqiang},
  journal={arXiv preprint arXiv:2504.19614},
  year={2025}
}

@inproceedings{
chen2025enhancing,
title={Enhancing {GUI} Agent with Uncertainty-Aware Self-Trained Evaluator},
author={Gongwei Chen and Lirong Jie and Lexiao Zou and Weili Guan and Miao Zhang and Liqiang Nie},
booktitle={The Thirty-ninth Annual Conference on Neural Information Processing Systems},
year={2025},
url={https://openreview.net/forum?id=cuUsD5FJbe}
}

@article{chen2020scene,
  title={Scene recognition with prototype-agnostic scene layout},
  author={Chen, Gongwei and Song, Xinhang and Zeng, Haitao and Jiang, Shuqiang},
  journal={IEEE Transactions on Image Processing},
  volume={29},
  pages={5877--5888},
  year={2020},
  publisher={IEEE}
}

@inproceedings{zou2025enhancing,
  title={Enhancing Diffusion-based Dataset Distillation via Adversary-Guided Curriculum Sampling},
  author={Zou, Lexiao and Chen, Gongwei and Chen, Yanda and Zhang, Miao},
  booktitle={2025 IEEE International Conference on Multimedia and Expo (ICME)},
  pages={1--6},
  year={2025},
  organization={IEEE}
}

@inproceedings{shao2017deep,
  title={Deep convolutional dynamic texture learning with adaptive channel-discriminability for 3D mask face anti-spoofing},
  author={Shao, Rui and Lan, Xiangyuan and Yuen, Pong C},
  booktitle={2017 IEEE International Joint Conference on Biometrics (IJCB)},
  pages={748--755},
  year={2017},
  organization={IEEE}
}

@article{lyu2026personalalign,
  title={PersonalAlign: Hierarchical Implicit Intent Alignment for Personalized GUI Agent with Long-Term User-Centric Records},
  author={Lyu, Yibo and Chen, Gongwei and Shao, Rui and Guan, Weili and Nie, Liqiang},
  journal={arXiv preprint arXiv:2601.09636},
  year={2026}
}

@article{shao2024detecting,
  title={Detecting and grounding multi-modal media manipulation and beyond},
  author={Shao, Rui and Wu, Tianxing and Wu, Jianlong and Nie, Liqiang and Liu, Ziwei},
  journal={IEEE Transactions on Pattern Analysis and Machine Intelligence},
  year={2024},
}

@inproceedings{chen2024spa,
  title={Spa-bench: A comprehensive benchmark for smartphone agent evaluation},
  author={Chen, Jingxuan and Yuen, Derek and Xie, Bin and Yang, Yuhao and Chen, Gongwei and Wu, Zhihao and Yixing, Li and Zhou, Xurui and Liu, Weiwen and Wang, Shuai and others},
  booktitle={NeurIPS 2024 Workshop on Open-World Agents},
  year={2024}
}

@inproceedings{shao2023detecting,
  title={Detecting and grounding multi-modal media manipulation},
  author={Shao, Rui and Wu, Tianxing and Liu, Ziwei},
  booktitle={Proceedings of the IEEE/CVF Conference on Computer Vision and Pattern Recognition},
  pages={6904--6913},
  year={2023}
}

@inproceedings{shao2019multi,
  title={Multi-adversarial discriminative deep domain generalization for face presentation attack detection},
  author={Shao, Rui and Lan, Xiangyuan and Li, Jiawei and Yuen, Pong C},
  booktitle={Proceedings of the IEEE/CVF conference on computer vision and pattern recognition},
  pages={10023--10031},
  year={2019}
}

@inproceedings{zhang2025cross,
  title={Cross-modal information flow in multimodal large language models},
  author={Zhang, Zhi and Yadav, Srishti and Han, Fengze and Shutova, Ekaterina},
  booktitle={Proceedings of the Computer Vision and Pattern Recognition Conference},
  pages={19781--19791},
  year={2025}
}

@article{li2025cogvla,
  title={Cogvla: Cognition-aligned vision-language-action model via instruction-driven routing \& sparsification},
  author={Li, Wei and Zhang, Renshan and Shao, Rui and He, Jie and Nie, Liqiang},
  journal={arXiv preprint arXiv:2508.21046},
  year={2025}
}

@article{zhu2025h,
  title={H-GAR: A Hierarchical Interaction Framework via Goal-Driven Observation-Action Refinement for Robotic Manipulation},
  author={Zhu, Yijie and Shao, Rui and Liu, Ziyang and He, Jie and Liu, Jizhihui and Wang, Jiuru and Yu, Zitong},
  journal={arXiv preprint arXiv:2511.17079},
  year={2025}
}

@article{li2025semanticvla,
  title={SemanticVLA: Semantic-Aligned Sparsification and Enhancement for Efficient Robotic Manipulation},
  author={Li, Wei and Zhang, Renshan and Shao, Rui and Fang, Zhijian and Zhou, Kaiwen and Tian, Zhuotao and Nie, Liqiang},
  journal={arXiv preprint arXiv:2511.10518},
  year={2025}
}

@article{zhang2025progrm,
  title={ProgRM: Build Better GUI Agents with Progress Rewards},
  author={Zhang, Danyang and Zhang, Situo and Yang, Ziyue and Zhu, Zichen and Zhao, Zihan and Cao, Ruisheng and Chen, Lu and Yu, Kai},
  journal={arXiv preprint arXiv:2505.18121},
  year={2025}
}

@article{luo2025navimaster,
  title={NaviMaster: Learning a Unified Policy for GUI and Embodied Navigation Tasks},
  author={Luo, Zhihao and Gong, Wentao Yan abd Jingyu and Wang, Min and Zhang, Zhizhong and Wang, Xuhong and Xie, Yuan and Tan, Xin},
  journal={arXiv preprint arXiv:2508.02046},
  year={2025}
}

@article{chen2025less,
  title={Less is More: Empowering GUI Agent with Context-Aware Simplification},
  author={Chen, Gongwei and Zhou, Xurui and Shao, Rui and Lyu, Yibo and Zhou, Kaiwen and Wang, Shuai and Li, Wentao and Li, Yinchuan and Qi, Zhongang and Nie, Liqiang},
  journal={arXiv preprint arXiv:2507.03730},
  year={2025}
}

@article{liu2025infigui,
  title={Infigui-r1: Advancing multimodal gui agents from reactive actors to deliberative reasoners},
  author={Liu, Yuhang and Li, Pengxiang and Xie, Congkai and Hu, Xavier and Han, Xiaotian and Zhang, Shengyu and Yang, Hongxia and Wu, Fei},
  journal={arXiv preprint arXiv:2504.14239},
  year={2025}
}

@article{rawles2023androidinthewild,
  title={Androidinthewild: A large-scale dataset for android device control},
  author={Rawles, Christopher and Li, Alice and Rodriguez, Daniel and Riva, Oriana and Lillicrap, Timothy},
  journal={Advances in Neural Information Processing Systems},
  volume={36},
  pages={59708--59728},
  year={2023}
}

@article{gao2025uishift,
  title={UIShift: Enhancing VLM-based GUI Agents through Self-supervised Reinforcement Learning},
  author={Gao, Longxi and Zhang, Li and Xu, Mengwei},
  journal={arXiv preprint arXiv:2505.12493},
  year={2025}
}

@article{li2024effects,
  title={On the effects of data scale on ui control agents},
  author={Li, Wei and Bishop, William E and Li, Alice and Rawles, Christopher and Campbell-Ajala, Folawiyo and Tyamagundlu, Divya and Riva, Oriana},
  journal={Advances in Neural Information Processing Systems},
  volume={37},
  pages={92130--92154},
  year={2024}
}

@inproceedings{chen2019transferability,
  title={Transferability vs. discriminability: Batch spectral penalization for adversarial domain adaptation},
  author={Chen, Xinyang and Wang, Sinan and Long, Mingsheng and Wang, Jianmin},
  booktitle={International conference on machine learning},
  pages={1081--1090},
  year={2019},
  organization={PMLR}
}

@inproceedings{xie2025gui,
title={GUI-explorer: Autonomous Exploration and Mining of Transition-aware Knowledge for GUI Agent},
author={Bin Xie and Rui Shao and Gongwei Chen and Kaiwen Zhou and Yinchuan Li and Jie Liu and Min Zhang and Liqiang Nie},
booktitle={Annual Meeting of the Association for Computational Linguistics (ACL)},
year={2025}
}

@article{achiam2023gpt,
  title={Gpt-4 technical report},
  author={Achiam, Josh and Adler, Steven and Agarwal, Sandhini and Ahmad, Lama and Akkaya, Ilge and Aleman, Florencia Leoni and Almeida, Diogo and Altenschmidt, Janko and Altman, Sam and Anadkat, Shyamal and others},
  journal={arXiv preprint arXiv:2303.08774},
  year={2023}
}

@article{hurst2024gpt,
  title={Gpt-4o system card},
  author={Hurst, Aaron and Lerer, Adam and Goucher, Adam P and Perelman, Adam and Ramesh, Aditya and Clark, Aidan and Ostrow, AJ and Welihinda, Akila and Hayes, Alan and Radford, Alec and others},
  journal={arXiv preprint arXiv:2410.21276},
  year={2024}
}

@inproceedings{chen2024lion,
  title={Lion: Empowering multimodal large language model with dual-level visual knowledge},
  author={Chen, Gongwei and Shen, Leyang and Shao, Rui and Deng, Xiang and Nie, Liqiang},
  booktitle={Proceedings of the IEEE/CVF Conference on Computer Vision and Pattern Recognition},
  pages={26540--26550},
  year={2024}
}

@InProceedings{zhang2025falcon,
    author={Zhang, Renshan and Shao, Rui and Chen, Gongwei and Zhang, Miao and Zhou, Kaiwen and Guan, Weili and Nie, Liqiang},
    title={FALCON: Resolving Visual Redundancy and Fragmentation in High-resolution Multimodal Large Language Models via Visual Registers},
    booktitle={Proceedings of the IEEE/CVF International Conference on Computer Vision (ICCV)},
    month= {October},
    year={2025},
}

@inproceedings{shen2024mome,
 author = {Shen, Leyang and Chen, Gongwei and Shao, Rui and Guan, Weili and Nie, Liqiang},
 booktitle = {Advances in Neural Information Processing Systems},
 pages = {42048--42070},
 title = {MoME: Mixture of Multimodal Experts for Generalist Multimodal Large Language Models},
 volume = {37},
 year = {2024}
}

@inproceedings{chen2024image,
  title={An image is worth 1/2 tokens after layer 2: Plug-and-play inference acceleration for large vision-language models},
  author={Chen, Liang and Zhao, Haozhe and Liu, Tianyu and Bai, Shuai and Lin, Junyang and Zhou, Chang and Chang, Baobao},
  booktitle={European Conference on Computer Vision},
  pages={19--35},
  year={2024},
  organization={Springer}
}

@inproceedings{lyu2025puma,
  title={PUMA: Layer-Pruned Language Model for Efficient Unified
Multimodal Retrieval with Modality-Adaptive Learning},
  author={Lyu, Yibo and Shao, Rui and Chen, Gongwei and Zhu, Yijie and Guan, Weili and Nie, Liqiang},
  booktitle={Proceedings of the 33nd ACM International Conference on Multimedia},
  year={2025}
}

@inproceedings{li2025lion,
  title={Lion-fs: Fast \& slow video-language thinker as online video assistant},
  author={Li, Wei and Hu, Bing and Shao, Rui and Shen, Leyang and Nie, Liqiang},
  booktitle={Proceedings of the Computer Vision and Pattern Recognition Conference},
  pages={3240--3251},
  year={2025}
}

@inproceedings{li2024optimus,
 author = {Li, Zaijing and Xie, Yuquan and Shao, Rui and Chen, Gongwei and Jiang, Dongmei and Nie, Liqiang},
 booktitle = {Advances in Neural Information Processing Systems},
 pages = {49881--49913},
 title = {Optimus-1: Hybrid Multimodal Memory Empowered Agents Excel in Long-Horizon Tasks},
 volume = {37},
 year = {2024}
}

@inproceedings{gou2024uground,
title={Navigating the Digital World as Humans Do: Universal Visual Grounding for {GUI} Agents},
author={Boyu Gou and Ruohan Wang and Boyuan Zheng and Yanan Xie and Cheng Chang and Yiheng Shu and Huan Sun and Yu Su},
booktitle={The Thirteenth International Conference on Learning Representations},
year={2025},
url={https://openreview.net/forum?id=kxnoqaisCT}
}

@article{luo2025gui,
  title={Gui-r1: A generalist r1-style vision-language action model for gui agents},
  author={Luo, Run and Wang, Lu and He, Wanwei and Xia, Xiaobo},
  journal={arXiv preprint arXiv:2504.10458},
  year={2025}
}

@article{lian2025ui,
  title={UI-AGILE: Advancing GUI Agents with Effective Reinforcement Learning and Precise Inference-Time Grounding},
  author={Lian, Shuquan and Wu, Yuhang and Ma, Jia and Song, Zihan and Chen, Bingqi and Zheng, Xiawu and Li, Hui},
  journal={arXiv preprint arXiv:2507.22025},
  year={2025}
}

@article{wang_mobile-agent_2024,
	title = {Mobile-{Agent}: {Autonomous} {Multi}-{Modal} {Mobile} {Device} {Agent} with {Visual} {Perception}},
	shorttitle = {Mobile-{Agent}},
	url = {http://arxiv.org/abs/2401.16158},
	doi = {10.48550/arXiv.2401.16158},
	urldate = {2024-05-29},
	journal = {arXiv preprint},
	author = {Wang, Junyang and Xu, Haiyang and Ye, Jiabo and Yan, Ming and Shen, Weizhou and Zhang, Ji and Huang, Fei and Sang, Jitao},
	month = apr,
	year = {2024},
	note = {arXiv:2401.16158 [cs]},
}

@article{lu_gui_2024,
	title = {{GUI} {Odyssey}: {A} {Comprehensive} {Dataset} for {Cross}-{App} {GUI} {Navigation} on {Mobile} {Devices}},
	shorttitle = {{GUI} {Odyssey}},
	url = {http://arxiv.org/abs/2406.08451},
	language = {en},
	urldate = {2024-06-13},
	journal = {arXiv preprint},
	author = {Lu, Quanfeng and Shao, Wenqi and Liu, Zitao and Meng, Fanqing and Li, Boxuan and Chen, Botong and Huang, Siyuan and Zhang, Kaipeng and Qiao, Yu and Luo, Ping},
	month = jun,
	year = {2024},
	note = {arXiv:2406.08451 [cs]},
}

@inproceedings{cheng_seeclick_2024,
	title = {{SeeClick}: {Harnessing} {GUI} {Grounding} for {Advanced} {Visual} {GUI} {Agents}},
	shorttitle = {{SeeClick}},
	url = {http://arxiv.org/abs/2401.10935},
	doi = {10.48550/arXiv.2401.10935},
	urldate = {2024-07-05},
	booktitle = {Proceedings of the 62nd {Annual} {Meeting} of the {Association} for {Computational} {Linguistics}},
	publisher = {arXiv},
	author = {Cheng, Kanzhi and Sun, Qiushi and Chu, Yougang and Xu, Fangzhi and Li, Yantao and Zhang, Jianbing and Wu, Zhiyong},
	month = feb,
	year = {2024},
	note = {arXiv:2401.10935 [cs]},
}

@misc{zhang_ui-hawk_2024,
	title = {{UI}-{Hawk}: {Unleashing} the {Screen} {Stream} {Understanding} for {GUI} {Agents}},
	shorttitle = {{UI}-{Hawk}},
	url = {https://www.preprints.org/manuscript/202408.2137/v1},
	doi = {10.20944/preprints202408.2137.v1},
	language = {en},
	urldate = {2024-09-09},
	publisher = {Preprints},
	author = {Zhang, Jiwen and Yu, Yaqi and Liao, Minghui and Li, Wentao and Wu, Jihao and Wei, Zhongyu},
	month = aug,
	year = {2024},
}

@inproceedings{lai_autowebglm_2024,
	title = {{AutoWebGLM}: {Bootstrap} {And} {Reinforce} {A} {Large} {Language} {Model}-based {Web} {Navigating} {Agent}},
	shorttitle = {{AutoWebGLM}},
	url = {http://arxiv.org/abs/2404.03648},
	doi = {10.48550/arXiv.2404.03648},
	urldate = {2024-09-11},
	booktitle = {Proceedings of the 30th {ACM} {SIGKDD} {Conference} on {Knowledge} {Discovery} and {Data} {Mining}},
	publisher = {arXiv},
	author = {Lai, Hanyu and Liu, Xiao and Iong, Iat Long and Yao, Shuntian and Chen, Yuxuan and Shen, Pengbo and Yu, Hao and Zhang, Hanchen and Zhang, Xiaohan and Dong, Yuxiao and Tang, Jie},
	month = apr,
	year = {2024},
	note = {arXiv:2404.03648 [cs]},
}

@article{chai_amex_2024,
	title = {{AMEX}: {Android} {Multi}-annotation {Expo} {Dataset} for {Mobile} {GUI} {Agents}},
	shorttitle = {{AMEX}},
	url = {http://arxiv.org/abs/2407.17490},
	doi = {10.48550/arXiv.2407.17490},
	urldate = {2024-09-18},
	journal = {arXiv preprint},
	author = {Chai, Yuxiang and Huang, Siyuan and Niu, Yazhe and Xiao, Han and Liu, Liang and Zhang, Dingyu and Gao, Peng and Ren, Shuai and Li, Hongsheng},
	month = jul,
	year = {2024},
	note = {arXiv:2407.17490 [cs]},
}

@article{wu_os-atlas_2024,
	title = {{OS}-{ATLAS}: {A} {Foundation} {Action} {Model} for {Generalist} {GUI} {Agents}},
	shorttitle = {{OS}-{ATLAS}},
	url = {http://arxiv.org/abs/2410.23218},
	doi = {10.48550/arXiv.2410.23218},
	urldate = {2024-11-06},
	journal = {arXiv preprint},
	author = {Wu, Zhiyong and Wu, Zhenyu and Xu, Fangzhi and Wang, Yian and Sun, Qiushi and Jia, Chengyou and Cheng, Kanzhi and Ding, Zichen and Chen, Liheng and Liang, Paul Pu and Qiao, Yu},
	month = oct,
	year = {2024},
	note = {arXiv:2410.23218 [cs]},
}

@article{xu_aguvis_2024,
	title = {Aguvis: {Unified} {Pure} {Vision} {Agents} for {Autonomous} {GUI} {Interaction}},
	shorttitle = {Aguvis},
	url = {http://arxiv.org/abs/2412.04454},
	language = {en},
	urldate = {2024-12-31},
	journal = {arXiv preprint},
	author = {Xu, Yiheng and Wang, Zekun and Wang, Junli and Lu, Dunjie and Xie, Tianbao and Saha, Amrita and Sahoo, Doyen and Yu, Tao and Xiong, Caiming},
	month = dec,
	year = {2024},
	note = {arXiv:2412.04454 [cs]},
}

@article{Schulman2017,
	title = {Proximal policy optimization algorithms},
	url = {http://arxiv.org/abs/1707.06347},
	author = {Schulman, John and Wolski, Filip and Dhariwal, Prafulla and Radford, Alec and Klimov, Oleg},
	year = {2017},
	note = {arXiv: 1707.06347},
	pages = {1--12},
}

@misc{luong_reft_2024,
	title = {{ReFT}: {Reasoning} with {Reinforced} {Fine}-{Tuning}},
	shorttitle = {{ReFT}},
	url = {http://arxiv.org/abs/2401.08967},
	language = {en},
	urldate = {2024-12-24},
	publisher = {arXiv},
	author = {Luong, Trung Quoc and Zhang, Xinbo and Jie, Zhanming and Sun, Peng and Jin, Xiaoran and Li, Hang},
	month = dec,
	year = {2024},
	note = {arXiv:2401.08967 [cs]},
}

@misc{liu_visual-rft_2025,
	title = {Visual-{RFT}: {Visual} {Reinforcement} {Fine}-{Tuning}},
	shorttitle = {Visual-{RFT}},
	url = {http://arxiv.org/abs/2503.01785},
	doi = {10.48550/arXiv.2503.01785},
	language = {en},
	urldate = {2025-03-10},
	publisher = {arXiv},
	author = {Liu, Ziyu and Sun, Zeyi and Zang, Yuhang and Dong, Xiaoyi and Cao, Yuhang and Duan, Haodong and Lin, Dahua and Wang, Jiaqi},
	month = mar,
	year = {2025},
	note = {arXiv:2503.01785 [cs]},
}

@misc{deepseek-ai_deepseek-r1_2025,
	title = {{DeepSeek}-{R1}: {Incentivizing} {Reasoning} {Capability} in {LLMs} via {Reinforcement} {Learning}},
	shorttitle = {{DeepSeek}-{R1}},
	url = {http://arxiv.org/abs/2501.12948},
	doi = {10.48550/arXiv.2501.12948},
	urldate = {2025-03-17},
	publisher = {arXiv},
	author = {DeepSeek-AI and Guo, Daya and Yang, Dejian and Zhang, Haowei and Song, Junxiao and Zhang, Ruoyu and Xu, Runxin and Zhu, Qihao and Ma, Shirong and Wang, Peiyi and Bi, Xiao and Zhang, Xiaokang and Yu, Xingkai and Wu, Yu and Wu, Z. F. and Gou, Zhibin and Shao, Zhihong and Li, Zhuoshu and Gao, Ziyi and Liu, Aixin and Xue, Bing and Wang, Bingxuan and Wu, Bochao and Feng, Bei and Lu, Chengda and Zhao, Chenggang and Deng, Chengqi and Zhang, Chenyu and Ruan, Chong and Dai, Damai and Chen, Deli and Ji, Dongjie and Li, Erhang and Lin, Fangyun and Dai, Fucong and Luo, Fuli and Hao, Guangbo and Chen, Guanting and Li, Guowei and Zhang, H. and Bao, Han and Xu, Hanwei and Wang, Haocheng and Ding, Honghui and Xin, Huajian and Gao, Huazuo and Qu, Hui and Li, Hui and Guo, Jianzhong and Li, Jiashi and Wang, Jiawei and Chen, Jingchang and Yuan, Jingyang and Qiu, Junjie and Li, Junlong and Cai, J. L. and Ni, Jiaqi and Liang, Jian and Chen, Jin and Dong, Kai and Hu, Kai and Gao, Kaige and Guan, Kang and Huang, Kexin and Yu, Kuai and Wang, Lean and Zhang, Lecong and Zhao, Liang and Wang, Litong and Zhang, Liyue and Xu, Lei and Xia, Leyi and Zhang, Mingchuan and Zhang, Minghua and Tang, Minghui and Li, Meng and Wang, Miaojun and Li, Mingming and Tian, Ning and Huang, Panpan and Zhang, Peng and Wang, Qiancheng and Chen, Qinyu and Du, Qiushi and Ge, Ruiqi and Zhang, Ruisong and Pan, Ruizhe and Wang, Runji and Chen, R. J. and Jin, R. L. and Chen, Ruyi and Lu, Shanghao and Zhou, Shangyan and Chen, Shanhuang and Ye, Shengfeng and Wang, Shiyu and Yu, Shuiping and Zhou, Shunfeng and Pan, Shuting and Li, S. S. and Zhou, Shuang and Wu, Shaoqing and Ye, Shengfeng and Yun, Tao and Pei, Tian and Sun, Tianyu and Wang, T. and Zeng, Wangding and Zhao, Wanjia and Liu, Wen and Liang, Wenfeng and Gao, Wenjun and Yu, Wenqin and Zhang, Wentao and Xiao, W. L. and An, Wei and Liu, Xiaodong and Wang, Xiaohan and Chen, Xiaokang and Nie, Xiaotao and Cheng, Xin and Liu, Xin and Xie, Xin and Liu, Xingchao and Yang, Xinyu and Li, Xinyuan and Su, Xuecheng and Lin, Xuheng and Li, X. Q. and Jin, Xiangyue and Shen, Xiaojin and Chen, Xiaosha and Sun, Xiaowen and Wang, Xiaoxiang and Song, Xinnan and Zhou, Xinyi and Wang, Xianzu and Shan, Xinxia and Li, Y. K. and Wang, Y. Q. and Wei, Y. X. and Zhang, Yang and Xu, Yanhong and Li, Yao and Zhao, Yao and Sun, Yaofeng and Wang, Yaohui and Yu, Yi and Zhang, Yichao and Shi, Yifan and Xiong, Yiliang and He, Ying and Piao, Yishi and Wang, Yisong and Tan, Yixuan and Ma, Yiyang and Liu, Yiyuan and Guo, Yongqiang and Ou, Yuan and Wang, Yuduan and Gong, Yue and Zou, Yuheng and He, Yujia and Xiong, Yunfan and Luo, Yuxiang and You, Yuxiang and Liu, Yuxuan and Zhou, Yuyang and Zhu, Y. X. and Xu, Yanhong and Huang, Yanping and Li, Yaohui and Zheng, Yi and Zhu, Yuchen and Ma, Yunxian and Tang, Ying and Zha, Yukun and Yan, Yuting and Ren, Z. Z. and Ren, Zehui and Sha, Zhangli and Fu, Zhe and Xu, Zhean and Xie, Zhenda and Zhang, Zhengyan and Hao, Zhewen and Ma, Zhicheng and Yan, Zhigang and Wu, Zhiyu and Gu, Zihui and Zhu, Zijia and Liu, Zijun and Li, Zilin and Xie, Ziwei and Song, Ziyang and Pan, Zizheng and Huang, Zhen and Xu, Zhipeng and Zhang, Zhongyu and Zhang, Zhen},
	month = jan,
	year = {2025},
	note = {arXiv:2501.12948 [cs]},
}

@misc{shao_deepseekmath_2024,
	title = {{DeepSeekMath}: {Pushing} the {Limits} of {Mathematical} {Reasoning} in {Open} {Language} {Models}},
	shorttitle = {{DeepSeekMath}},
	url = {http://arxiv.org/abs/2402.03300},
	doi = {10.48550/arXiv.2402.03300},
	urldate = {2025-03-17},
	publisher = {arXiv},
	author = {Shao, Zhihong and Wang, Peiyi and Zhu, Qihao and Xu, Runxin and Song, Junxiao and Bi, Xiao and Zhang, Haowei and Zhang, Mingchuan and Li, Y. K. and Wu, Y. and Guo, Daya},
	month = apr,
	year = {2024},
	note = {arXiv:2402.03300 [cs]},
}

@article{UI-hawk,
	doi = {10.20944/preprints202408.2137.v1},
	url = {https://doi.org/10.20944/preprints202408.2137.v1},
	year = 2024,
	month = {August},
	publisher = {Preprints},
	author = {Jiwen Zhang and Yaqi Yu and Minghui Liao and Wentao Li and Jihao Wu and Zhongyu Wei},
	title = {UI-Hawk: Unleashing the Screen Stream Understanding for GUI Agents},
	journal = {Preprints}
}

@misc{lu_ui-r1_2025,
	title = {{UI}-{R1}: {Enhancing} {Action} {Prediction} of {GUI} {Agents} by {Reinforcement} {Learning}},
	shorttitle = {{UI}-{R1}},
	url = {http://arxiv.org/abs/2503.21620},
	doi = {10.48550/arXiv.2503.21620},
	urldate = {2025-03-28},
	publisher = {arXiv},
	author = {Lu, Zhengxi and Chai, Yuxiang and Guo, Yaxuan and Yin, Xi and Liu, Liang and Wang, Hao and Xiong, Guanjing and Li, Hongsheng},
	month = mar,
	year = {2025},
	note = {arXiv:2503.21620 [cs]
version: 1},
}

@misc{tan_reason-rft_2025,
	title = {Reason-{RFT}: {Reinforcement} {Fine}-{Tuning} for {Visual} {Reasoning}},
	shorttitle = {Reason-{RFT}},
	url = {http://arxiv.org/abs/2503.20752},
	doi = {10.48550/arXiv.2503.20752},
	urldate = {2025-04-01},
	publisher = {arXiv},
	author = {Tan, Huajie and Ji, Yuheng and Hao, Xiaoshuai and Lin, Minglan and Wang, Pengwei and Wang, Zhongyuan and Zhang, Shanghang},
	month = mar,
	year = {2025},
	note = {arXiv:2503.20752 [cs]},
}
}

\clearpage
\setcounter{page}{1}

\begin{figure*}[t]
    \centering
    \includegraphics[width=0.75\textwidth]{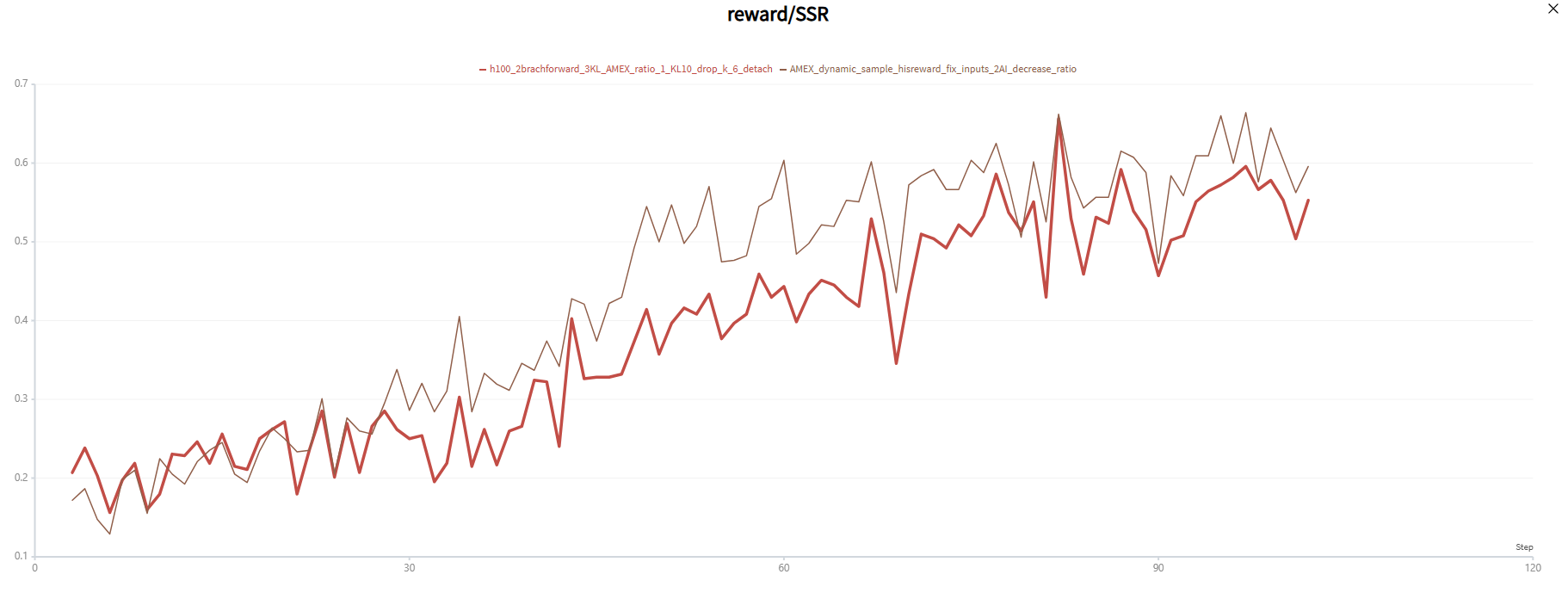}
\caption{Training accuracy curves of Hicon-Agent with and without DCS under the AHC framework. Models trained with DCS exhibit consistently higher accuracy and faster convergence, demonstrating that adaptive history sampling facilitates more effective learning.}
    \label{fig:train_curve}
\vspace{-3mm}
\end{figure*}


\section*{A. Visualization of SSR curve during training}
During training, we monitor the execution accuracy to evaluate the learning behavior of the policy over time. As shown in Figure ~\ref{fig:train_curve}, integrating DCS into the AHC framework leads to a clear improvement in training dynamics. The curve with DCS achieves both higher convergence speed and better final accuracy compared to the AHC-only baseline. This indicates that naively relying on fixed-length histories limits the model's ability to generalize. In contrast, DCS adaptively determines the effective history length based on task complexity, enabling the agent to access relevant information while avoiding overfitting to noisy or irrelevant historical states. These results verify that dynamically sampled history promotes more stable optimization and enhances the agent's capability to learn meaningful sequential dependencies.

\section*{B. Visualization of sampling distribution in DCS.}
\begin{figure}[H]
    \centering
    \includegraphics[width=0.47\textwidth]{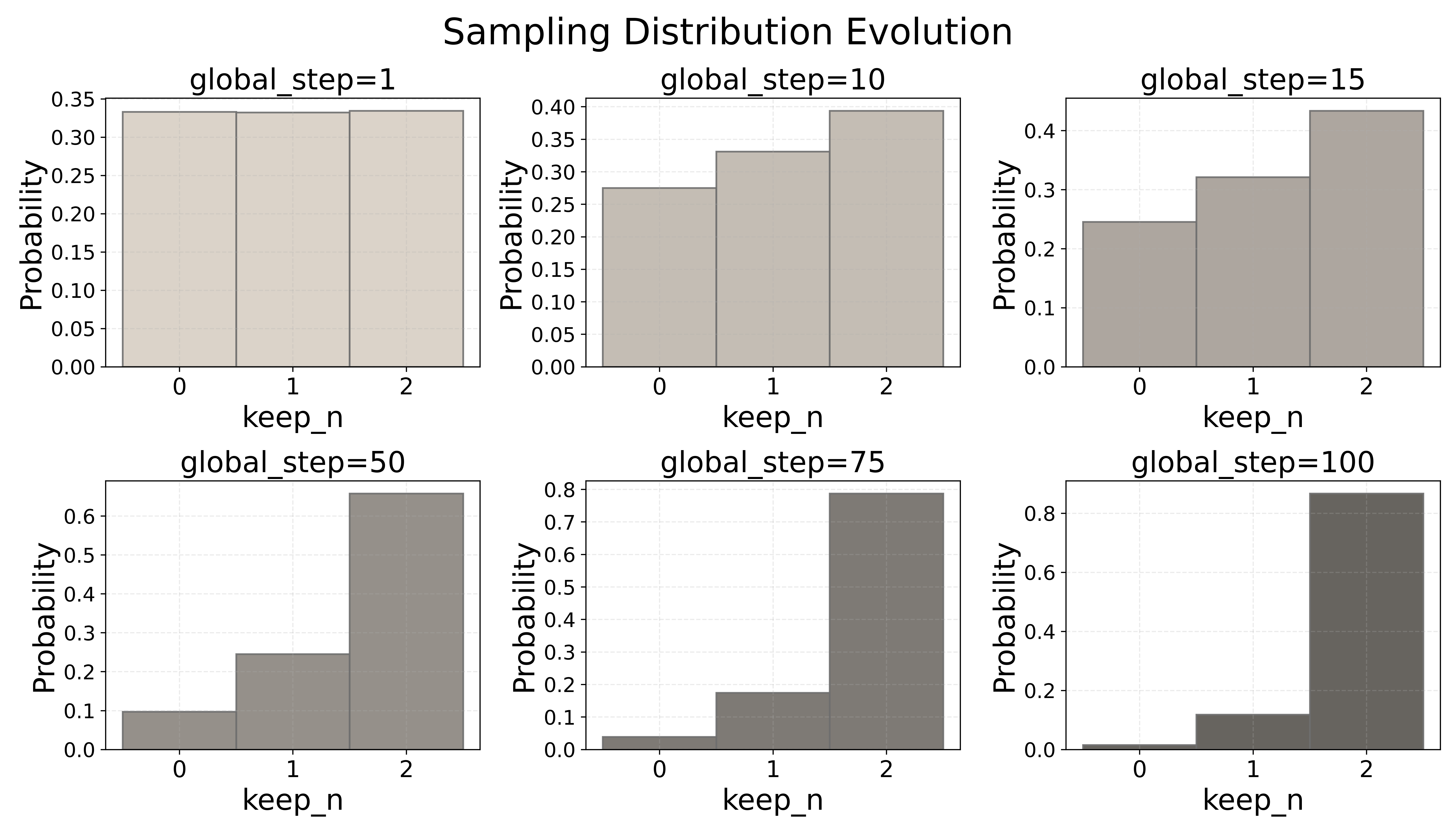}
\caption{Evolution of the sampling distribution from uniform exploration to exponentially biased selection as training progresses.}
    \label{fig:distribution_evolve}
\vspace{-3mm}
\end{figure}
We sample the number of preserved history blocks $\tau_i \in \{0,1,2\}$ from a time-dependent exponential distribution. At training step $u$, the sampling distribution is defined as
\[
P(\tau_i \mid u) = \frac{\exp\big(\lambda(u)\, \tau_i\big)}{\sum_{j=0}^{N} \exp\big(\lambda(u)\, j\big)},
\]
where $\lambda(u)$ is a linear function that increases with $u$. As shown in Figure~\ref{fig:distribution_evolve}, in the early stage of training, $\lambda(u)\approx 0$ and the distribution is nearly uniform, encouraging random exploration. As training progresses, $\lambda(u)$ gradually grows, yielding an increasingly biased exponential distribution that favors larger values of $\tau_i$. This schedule smoothly shifts the sampling strategy from random selection to full-context history.

We define $\lambda(u)$ as a function that increases from $0$ to $\lambda_{\max}$ within the first $\alpha T$ steps, gradually shifting the sampling distribution from uniform to exponentially biased:
\begin{equation}
    \lambda(u) = \lambda_{\max} \cdot \min\left(1,\; \frac{u}{\alpha T}\right)
\end{equation}
$T$ denotes the total number of training steps. $\lambda_{\max}$ controls the final steepness of the exponential bias (larger $\lambda_{\max}$ yields a more strongly peaked distribution that favors larger $\tau_i$), while $\alpha$ determines the fraction of training used for warm-up. In our implementation, we set $\lambda_{\max}=2$, $\alpha=\tfrac{1}{3}$.

\section*{C. GUI datasets}
\label{appendix:dataset}
We construct our training data from the open-source AMEX dataset\citep{chai_amex_2024}, which contains high-level GUI interaction trajectories. Preserving the original action distribution, we randomly sample 3,000 steps without applying additional filtering or cleaning procedures.

\paragraph{AMEX}
AMEX is a large-scale mobile GUI dataset for training and evaluating control agents, comprising $>$104K high-resolution screenshots from 110 Android apps with multi-level annotations. Each episode contains about $13$ actions on average.

\textit{Action distribution (full dataset ratio):}
\begin{itemize}\setlength{\itemsep}{1pt}
\item \texttt{click(start\_box='(x,y)')} \textbf{24815 (64.11\%)}
\item \texttt{scroll(direction='down or up or right or left')} \textbf{7628 (19.71\%)}
\item \texttt{finished()} \textbf{2828 (7.31\%)}
\item \texttt{type(content='')} \textbf{2419 (6.25\%)}
\item \texttt{press\_enter()} \textbf{651 (1.68\%)}
\item \texttt{impossible()} \textbf{220 (0.57\%)}
\item \texttt{press\_back()} \textbf{135 (0.35\%)}
\item \texttt{press\_home()} \textbf{13 (0.03\%)}
\end{itemize}

\paragraph{AndroidControl}
AndroidControl is diverse benchmark to study data scaling for UI control, containing 15{,}283 demonstrations spanning 14{,}548 unique tasks across 833 Android apps, with both high-level and low-level human-written instructions for each task. Each episode contains about $5$ actions on average.

\textit{Action distribution (test split):}
\begin{itemize}\setlength{\itemsep}{1pt}
\item \texttt{click(start\_box='(x,y)')} \textbf{5074 (50.81\%)}
\item \texttt{finished()} \textbf{1543 (15.45\%)}
\item \texttt{scroll(direction='down or up or right or left')} \textbf{1211 (12.13\%)}
\item \texttt{type(content='')} \textbf{632 (6.33\%)}
\item \texttt{open\_app(app\_name='')} \textbf{608 (6.09\%)}
\item \texttt{wait()} \textbf{567 (5.68\%)}
\item \texttt{press\_back()} \textbf{343 (3.43\%)}
\item \texttt{long\_press(start\_box='(x,y)')} \textbf{9 (0.09\%)}
\end{itemize}

\paragraph{GUI-Odyssey}
GUI-Odyssey is a cross-app mobile GUI navigation dataset for multi-step workflows across apps; the paper reports 7{,}735 episodes over 6 devices, 6 task types, 201 apps and $\sim$1.4K app combinations. Each episode contains about $15 $ actions on average.

\textit{Action distribution (test split):}
\begin{itemize}\setlength{\itemsep}{1pt}
\item \texttt{click(start\_box='(x,y)')} \textbf{19142 (65.05\%)}
\item \texttt{type(content='')} \textbf{3113 (10.58\%)}
\item \texttt{scroll(direction='down or up or right or left')} \textbf{2764 (9.39\%)}
\item \texttt{press\_home()} \textbf{2233 (7.59\%)}
\item \texttt{finished()} \textbf{1875 (6.37\%)}
\item \texttt{long\_press(start\_box='(x,y)')} \textbf{106 (0.36\%)}
\item \texttt{press\_recent()} \textbf{74 (0.25\%)}
\item \texttt{press\_back()} \textbf{61 (0.21\%)}
\item \texttt{impossible()} \textbf{58 (0.20\%)}
\end{itemize}

\paragraph{AITW}
Android in the Wild (AITW) is a large-scale dataset for Android device control using natural language instructions. The paper reports ~715,000 episodes, 30,000+ unique instructions, 8 device types (Pixel 2 XL through Pixel 6), 4 Android versions, covering hundreds of apps and websites. Each episode contains about 6.5 actions on average.

\section*{D. Effect of history observation and compression}
We first examine the impact of historical observations by comparing Qwen2.5VL-3B(2A), which uses only past actions, with Qwen2.5VL-3B(2AO), which includes both actions and observations from the past two steps. Incorporating visual history brings clear improvements across all metrics (+8.96\% SR), highlighting the importance of visual context in guiding decision-making. When applying inference-only compression to the 2AO model, performance drops significantly (-4.95\% SR). In contrast, our \textbf{Hicon-Agent-3B}, trained with history-aware optimization, recovers most of this loss and improves performance(+5.06 \% SR) compared to the compressed baseline. It also exceeds the uncompressed 2AO baseline in SR and grounding accuracy, demonstrating more effective and efficient usage of historical context. This highlights the advantage of our training strategy in mitigating the trade-off between computational efficiency and task performance.
\begin{table}[h]
    \centering
    \footnotesize
    \setlength{\tabcolsep}{0.8pt} 
    \renewcommand{\arraystretch}{1.3} 
    \begin{tabular}{l|cccccc}
    \toprule[1pt]
    Models & Hist. & Comp. & Type & Ground. & SR & FLOPs \\
    \midrule
    Qwen2.5VL-3B & 2A  & -- & 61.05 & 59.61 & 43.33 & 13.21 \\
    Qwen2.5VL-3B & 2AO & -- & 69.29 & 63.88 & 52.29 & 35.75 \\
    Qwen2.5VL-3B & 2AO & \checkmark & 
        66.56{\tiny\textcolor{drop}{(-2.73)}} & 
        61.10{\tiny\textcolor{drop}{(-2.78)}} & 
        47.34{\tiny\textcolor{drop}{(-4.95)}} & 25.21{\tiny\textcolor{improve}{(-10.54)}} \\
    \rowcolor{rowgray}
    Hicon-Agent-3B & 2AO & \checkmark & 
        67.79{\tiny\textcolor{improve}{(+1.23)}} & 
        65.01{\tiny\textcolor{improve}{(+3.91)}} & 
        52.40{\tiny\textcolor{improve}{(+5.06)}} & 25.21{\tiny\textcolor{improve}{(-10.54)}} \\
    \bottomrule[1pt]
    \end{tabular}
    \caption{Study on different strategies for history utilization on the AndroidControl dataset. \textcolor{improve}{Red} indicates improvement, \textcolor{drop}{green} indicates drop. \textbf{Hist.} denotes the history context format, and \textbf{Comp.} indicates whether history compression is applied.}
    \label{tab:amex_ablation}
    \vspace{-6mm}
\end{table}

\section*{E. Algorithm details}
Algorithm~\ref{alg:hcpo} details the HCPO training loop. We begin with on-policy, group-wise rollouts using Dynamic Context Sampling: for each of the $G$ samples, we first construct the full input context $\{I, H_t, s_t\}$, then draw a truncated history $H_t^{i}$ by sampling a history length $\tau_i \leq \tau$ sampled from the distribution $p$.  Given this truncated context, we sample a response $o_i \sim \pi_{\theta_{\text{old}}}(\cdot \mid I, H_t^{i}, s_t)$. The corresponding reward $\{r_i\}$ is computed and converted into group advantages $\{A_i\}$. 

Each sampled response is then evaluated by two forward passes of the current policy with shared parameters: (i) an uncompressed branch that uses $\{I, H_t, s_t\}$ end-to-end; and (ii) a compressed branch that mirrors the first $k$ layers and then drops history \emph{vision} tokens after layer $k$, retaining action and other tokens to form $H_t^{i,c}$. Reusing the same $\{o_i\}$ isolates the effect of compression. 

We optimize two clipped-ratio gradient policy losses against $\pi_{\theta}$, while adding a token-level consistency term that pulls the compressed distribution $\pi_{\theta}(\cdot \mid I, H_t^{i,c}, s_t)$ toward the uncompressed distribution $\pi_{\theta}(\cdot \mid I, H_t^i, s_t)$. Teacher logits from the uncompressed branch are detached to prevent gradient flow. The final objective $L_{\text{HCPO}}$ preserves on-policy learning under complete history and aligns the compressed path for efficient inference.

\begin{algorithm}
\caption{History Context-aware Policy Optimization (HCPO)}
\label{alg:hcpo}
\begin{algorithmic}[1]
\Require Policy model $\pi_\theta$, old policy $\pi_{\theta_{\text{old}}}$, reward model $R$, task instruction $I$, current GUI observation $s_t$, history context $H_t = \{(s_{t-\tau}, a_{t-\tau}), \dots, (s_{t-1}, a_{t-1})\}$, group size $G$, compression layer $k$, consistency weight $\lambda$
\vspace{0.5ex}

\State \textcolor{orange}{\# Group rollout with DCS}
\State Initialize sampling distribution $p$
\State Build full history sequence: $\{I, H_t, s_t\}$
\For{$i = 1$ to $G$}
    \State Sample history length $\tau_i\sim p$ to get $H_t^i$
    \State Sample and select response $o_i \sim \pi_{\theta_{\text{old}}}(\cdot \mid I, H_t^i, s_t)$
\EndFor
\State Evaluate rewards $\{r_1, \dots, r_G\} \leftarrow R(\{o_1, \dots, o_G\})$
\State Compute group-wise advantages $\{A_1, \dots, A_G\}$

\State \textcolor{orange}{\# Full history branch forward pass}
\For{$i = 1$ to $G$}
    \State Compute logits $\pi_{\theta_{old}}(o_i \mid I, H_t^i, s_t))$
    \State Compute logits $\pi_\theta(o_i \mid I, H_t^i, s_t))$
    \State Compute reference logits $\pi_{\theta_{\text{ref}}}(o_i \mid I, H_t^i, s_t)$
\EndFor

\vspace{0.5ex}
\State \textcolor{orange}{\# Compressed history branch forward pass}
\For{$i = 1$ to $G$}
    \State Reuse response $o_i$ and perform forward pass with partial history compression:
    \State \hspace{1em} Use uncompressed input $\{I, H_t^i, s_t\}$ in first $k$ layers
    \State \hspace{1em} Drop history vision tokens from $H_t^i$ in layers $k+1$ to get $H_t^{i,c}$
    \State Compute compressed logits $\pi_\theta(o_i \mid I, H_t^{i,c}, s_t)$
    \State Compute token-level KL divergence: $\mathbb{D}_{\text{KL}}(\pi_\theta(o_i \mid I, H_t^{i,c}, s_t) \parallel \pi_\theta(o_i \mid I, H_t^i, s_t))$
\EndFor
\State \textcolor{orange}{\# Compute policy-gradient loss and consistency loss}

\State Compute uncompressed policy loss $\mathcal{L}_{\text{w/o comp}}$
\State Compute compressed policy loss $\mathcal{L}_{\text{w/ comp}}$ 
\State Compute consistency loss: $\mathcal{L}_{\text{KL}} = \sum_{i=1}^G \mathbb{D}_{\text{KL}}(\cdot)$

\vspace{0.5ex}
\State Compute total loss: $\mathcal{L}_{\text{HCPO}} = \mathcal{L}_{\text{w/o comp}} + \mathcal{L}_{\text{w/ comp}} + \lambda \mathcal{L}_{\text{KL}}$
\State Update model: $\theta \leftarrow \theta - \eta \nabla_\theta \mathcal{L}_{\text{HCPO}}$
\end{algorithmic}
\end{algorithm}

\section*{F. Prompts for training and evaluation}
\definecolor{promptBack}{HTML}{F5F7F9}
\definecolor{promptFrame}{HTML}{D6DCE5}
\definecolor{promptTag}{HTML}{B04A6B}
\definecolor{promptComment}{HTML}{2E7D7D}
\definecolor{promptText}{HTML}{2B2B2B}
\definecolor{keywordgreen}{rgb}{0,0.6,0}

\lstdefinestyle{llmPrompt}{
  backgroundcolor=\color{promptBack},
  basicstyle=\ttfamily\small\color{promptText},
  columns=fullflexible, keepspaces=true, showstringspaces=false,
  breaklines=true,  
  breakatwhitespace=false,       
  aboveskip=6pt, belowskip=6pt,
  xleftmargin=1.2em, xrightmargin=1.2em,
  frame=single, framerule=0.4pt, framesep=6pt,
  rulecolor=\color{promptFrame},
  captionpos=b,
  breakindent=0pt,
  morecomment=[l][\itshape\color{promptComment}]{\#\#},
  moredelim=[s][\bfseries\color{promptTag}]{<}{>},
  emph={click,type,scroll,press_back,press_home,press_enter,finished},
  emphstyle=\bfseries\color{keywordgreen}
}

\begin{lstlisting}[style=llmPrompt,
  caption={AMEX training prompt template.},
  label={lst:amex_prompt}]
You are a GUI agent. You are given a task and your action history, with screenshots. You need to perform the next action to complete the task. You FIRST need to think based on the current image, task, and historical actions. The reasoning process MUST BE enclosed within <think> </think> tags. Then output the action, which MUST BE put in <action> </action> and MUST BE in Action Space. 

## Output Format
<think>...</think><action>...</action>

## Action Space
click(start_box='(x,y)')
type(content='')
scroll(direction='down or up or right or left')
press_back()
press_home()
press_enter()
finished()

## Example:
<think>The user wants to search for shoes. The current screen has a search bar at the top.</think>
<action>click(start_box='(x,y)')</action>
\end{lstlisting}

\begin{lstlisting}[style=llmPrompt,
  caption={AndroidControl evaluation prompt template.},
  label={lst:AC_prompt}]
You are a GUI agent. You are given a task and your action history, with screenshots. You need to perform the next action to complete the task. You FIRST need to think based on the current image, task, and historical actions. The reasoning process MUST BE enclosed within <think> </think> tags. Then output the action, which MUST BE put in <action> </action> and MUST BE in Action Space. 

## Output Format
<think>...</think><action>...</action>

## Action Space
click(start_box='(x,y)')
long_press(start_box='(x,y)')
type(content='')
scroll(direction='down or up or right or left')
open_app(app_name='')
press_back()
press_home()
wait()
finished()

## Example:
<think>The user wants to search for shoes. The current screen has a search bar at the top.</think>
<action>click(start_box='(x,y)')</action>
\end{lstlisting}

\begin{lstlisting}[style=llmPrompt,
  caption={GUI-Odyssey evaluation prompt template.},
  label={lst:gui-odyssey_prompt}]
You are a GUI agent. You are given a task and your action history, with screenshots. You need to perform the next action to complete the task. You FIRST need to think based on the current image, task, and historical actions. The reasoning process MUST BE enclosed within <think> </think> tags. Then output the action, which MUST BE put in <action> </action> and MUST BE in Action Space. 

## Output Format
<think>...</think><action>...</action>

## Action Space
click(start_box='(x,y)')
long_press(start_box='(x,y)')
type(content='')
scroll(direction='down or up or right or left')
impossible()
press_back()
press_home()
press_recent()
finished()
## Example:
<think>The user wants to search for shoes. The current screen has a search bar at the top.</think>
<action>click(start_box='(x,y)')</action>
\end{lstlisting}

\begin{lstlisting}[style=llmPrompt,
  caption={AITW evaluation prompt template.},
  label={lst:aitw_prompt}]
You are a GUI agent. You are given a task and your action history, with screenshots. You need to perform the next action to complete the task. You FIRST need to think based on the current image, task, and historical actions. The reasoning process MUST BE enclosed within <think> </think> tags. Then output the action, which MUST BE put in <action> </action> and MUST BE in Action Space. 

## Output Format
<think>...</think><action>...</action>

## Action Space
click(start_box='(x,y)')
long_press(start_box='(x,y)')
type(content='')
scroll(direction='down or up or right or left')
impossible()
press_enter()
press_back()
press_home()
finished()
## Example:
<think>The user wants to search for shoes. The current screen has a search bar at the top.</think>
<action>click(start_box='(x,y)')</action>
\end{lstlisting}


\section*{G. Model Behavior Across Different History Lengths}
\begin{figure*}[h]
    \centering
    \setlength{\abovecaptionskip}{0mm}
    \setlength{\belowcaptionskip}{0mm}
    \includegraphics[width=0.9\textwidth]{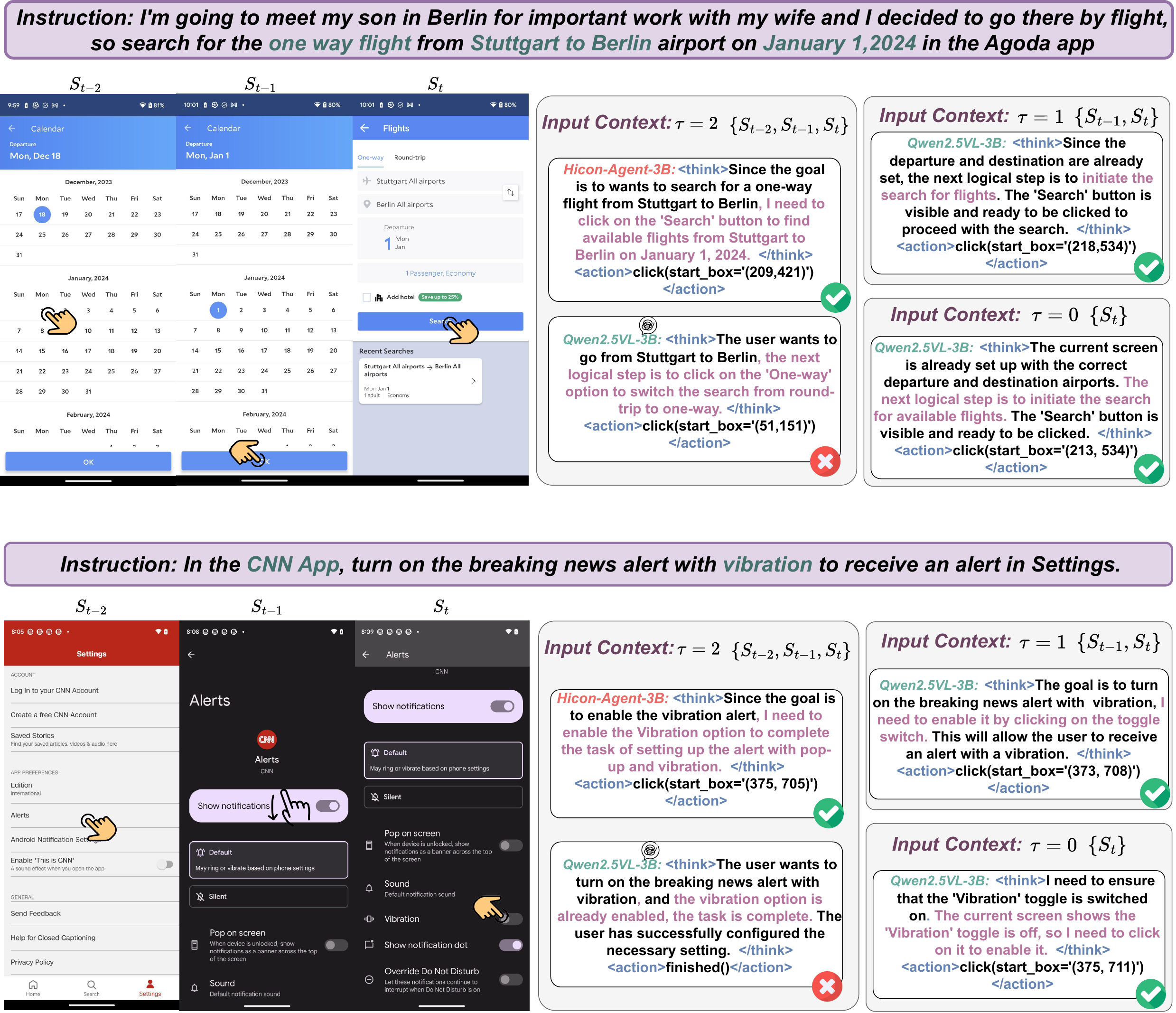}
\caption{Case study of model behavior under different history length context.}
    \label{fig:case2_study_his}
\vspace{0mm}
\end{figure*}

To better understand how history length affects agent behavior, we provide a case study comparing the base model and our HiconAgent-3B under different history lengths $\tau \in \{0,1,2\}$. As shown in Figure \ref{fig:case2_study_his}, the base model performs correctly when using shorter contexts ($\tau=0$ or $\tau=1$), but fails when the history is extended to $\tau=2$, where the additional observations introduce distracting or misleading information, causing the model to attend to an incorrect UI element and produce the wrong action. In contrast, our model, trained with Dynamic Context Sampling, still produces the correct action when $\tau=2$. Since DCS exposes the agent to diverse and progressively biased history lengths during optimization, the model learns to effectively utilize extended context. This qualitative evidence supports our quantitative results, demonstrating that naively increasing history is suboptimal, whereas HCPO equips the agent with robustness across variable context windows and enables it to benefit from longer history when necessary.

\section*{H. Visualization of successful trajectories}
\begin{figure*}[h]
    \centering
    \setlength{\abovecaptionskip}{0mm}
    \setlength{\belowcaptionskip}{0mm}

    \begin{subfigure}{0.48\textwidth}
        \centering
        \includegraphics[width=\textwidth]{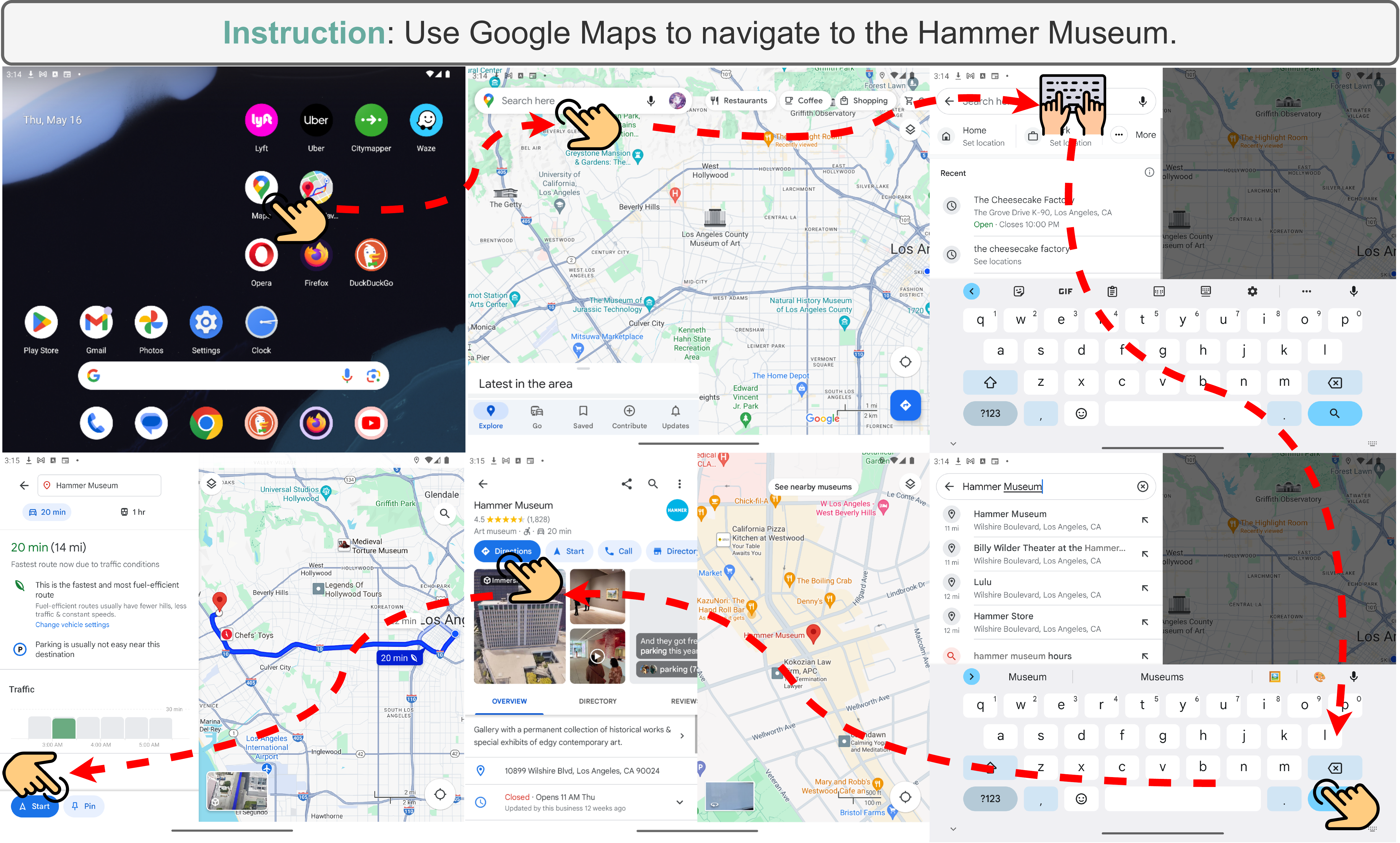}
        \caption{Trajectory example 1}
        \label{fig:case1_traj}
    \end{subfigure}
    \hfill
    \begin{subfigure}{0.48\textwidth}
        \centering
        \includegraphics[width=\textwidth]{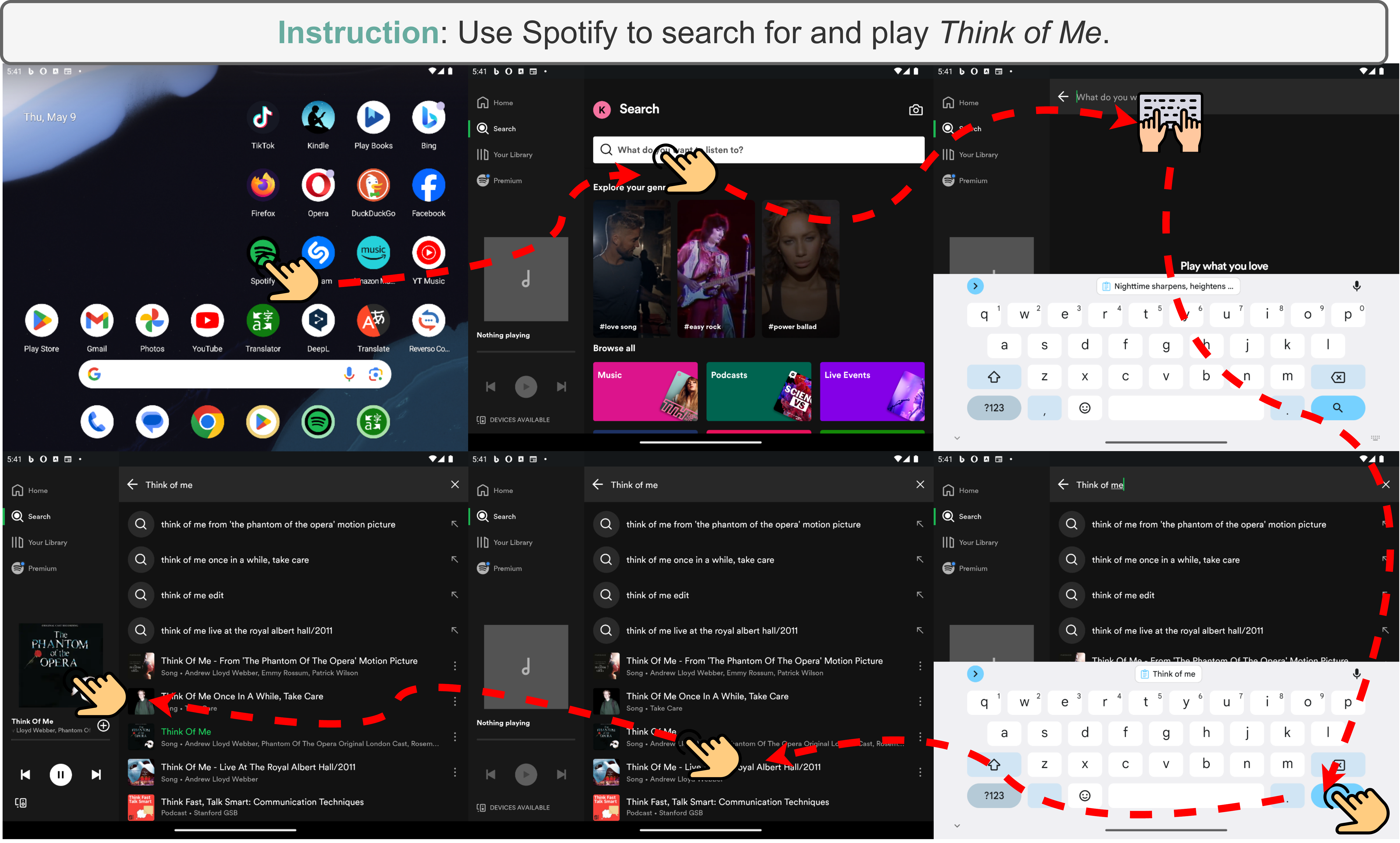}
        \caption{Trajectory example 2}
        \label{fig:case2_traj}
    \end{subfigure}

    \caption{Case studies of our model on downstream GUI navigation tasks.}
    \label{fig:case_study_traj}
    \vspace{0mm}
\end{figure*}

To better illustrate how our proposed HCPO framework facilitates robust decision-making, we visualize several representative successful trajectories from the evaluation benchmarks as shown in Figure \ref{fig:case1_traj} and Figure \ref{fig:case2_traj}.




\end{document}